\documentclass{article} 

\usepackage{PRIMEarxiv}

\usepackage[english]{babel}
\usepackage{amsfonts}
\usepackage{amsmath}
\usepackage{amsthm}
\usepackage[utf8]{inputenc}
\usepackage{graphicx}
\usepackage{todonotes}
\usepackage[ruled, linesnumbered, noend]{algorithm2e}
\usepackage{placeins}
\usepackage{subcaption}
\usepackage{rotating}
\usepackage{hyperref}
\usepackage{fancyhdr}
\usepackage{placeins}
\usepackage{float}
\usepackage{booktabs}
\usepackage{multirow}
\usepackage{orcidlink}
\title{Dynamically Weighted Federated k-Means}

\author{
  Patrick Holzer \orcidlink{0009-0004-2975-8365}, Tania Jacob \orcidlink{0009-0002-1404-0356}, Shubham Kavane \orcidlink{0009-0004-4715-1452}\\
  Fraunhofer Institute for Industrial Mathematics ITWM, Department for Financial Mathematics \\
  Fraunhofer-Platz 1, 67663 Kaiserslautern, Germany\\
  \texttt{\{patrick.holzer, tania.jacob, shubham.kavane\}@itwm.fraunhofer.de}
}

\newcommand\norm[1]{\lVert#1\rVert}

\newcommand{\R}{\mathbb{R}}

\begin{document}
\maketitle

% Abstract (Do not insert blank lines, i.e. \\) 
\begin{abstract}
Federated clustering, an integral aspect of federated machine learning, enables multiple data sources to collaboratively cluster their data, maintaining decentralization and preserving privacy.
In this paper, we introduce a novel federated clustering algorithm  named \texttt{Dynamically Weighted Federated k-means (DWF k-means)} based on Lloyd’s method for k-means clustering, to address the challenges associated with distributed data sources and heterogeneous data. 
Our proposed algorithm combines the benefits of traditional clustering techniques with the privacy and scalability benefits offered by federated learning.
The algorithm facilitates collaborative clustering among multiple data owners, allowing them to cluster their local data collectively while exchanging minimal information with the central coordinator. The algorithm optimizes the clustering process by adaptively aggregating cluster assignments and centroids from each data source, thereby learning a global clustering solution that reflects the collective knowledge of the entire federated network. We address the issue of empty clusters, which commonly arises in the context of federated clustering.
We conduct experiments on multiple datasets and data distribution settings to evaluate the performance of our algorithm in terms of clustering score, accuracy, and v-measure. 
The results demonstrate that our approach can match the performance of the centralized classical k-means baseline, and outperform existing federated clustering methods like \texttt{k-FED} in realistic scenarios.
\end{abstract}

% Keywords
%\keyword{keyword 1; keyword 2; keyword 3 (List three to ten pertinent keywords specific to the article; yet reasonably common within the subject discipline.)} 

\section{Introduction}
\label{sec:chapter_intro}
In today's era of unprecedented data proliferation, the analysis and utilization of large and complex datasets has become a driving force behind scientific discoveries, business insights, and technological advancements. With the growing importance of data-driven decision-making, it is necessary to devise innovative techniques that can effectively and efficiently process and extract knowledge from these vast, decentralized data sources. In the last decade, federated learning \cite{mcmahan2017communication} has emerged as a powerful paradigm for collaborative data analysis and machine learning in decentralized environments, allowing multiple parties to jointly train models without exchanging raw data. Rather than requiring the transmission of data to a central location, devices perform local training utilizing their own data and subsequently share only the essential model updates. While there has been a considerable amount of research into federated learning, its application to clustering tasks remains a challenging frontier.

Clustering is a fundamental machine learning technique used to group together points, based on their inherent similarities. It is an unsupervised learning approach that is useful in tasks such as data understanding, pattern discovery, dimension reduction, segmentation, personalization and anomaly detection \cite{rai2010survey}. It is often used to enable data-driven decisions and pave the way for the application of subsequent machine learning techniques. While centralized clustering techniques are well established, there have been fewer works considering distributed clustering, and fewer still that take into account the unique challenges of federated learning \cite{dennis2021heterogeneity, Wang2020FederatedCV, Li2022DifferentiallyPV, Wang2023FederatedFK}. 

One important challenge is the statistical heterogeneity arising from the diversity of data. This diversity may stem from varying data volumes and data distributions across the participating devices, due to factors such as differences in data collection processes, user behaviors, and local environments. Such non-independent and identically distributed (non-IID) data increase the complexity of the modeling process and are therefore sometimes ignored in distributed optimization, which is why IID data is often assumed there \cite{li2020federated}. Contrary to non-IID data, IID data implies that each data point in a dataset is independent of all others, and that all data points are drawn from the same probability distribution. This means that they share the same statistical properties and hence simplifies modeling tasks. 
However, making this assumption is not suitable for most real-world situations and can adversely impact the effectiveness and efficiency of federated learning methods. If left unaddressed, statistical heterogeneity can result in reduced model generalization, biased models, increased communication overhead and convergence time, as well as exacerbate privacy concerns \cite{li2020federated, yang2021characterizing, abdelmoniem2023comprehensive}. Therefore, in this work, we develop and analyze a \texttt{Dynamically Weighted Federated k-means (DWF k-means)} method, that focuses on addressing the challenge of statistical heterogeneity. 

This proposed method introduces dynamic weighting of the model updates from each participating client, allowing the algorithm to give more importance to certain updates and less importance to others, to account for heterogeneity when forming clusters. The dynamic nature of the weighting allows this importance to vary for each cluster and client per federated learning round, making it a more powerful technique than using constant weights over all rounds, which we name \texttt{Equally Weighted Federated k-means (EWF k-means)}. 

To evaluate the effectiveness of \texttt{DWF k-means}, we conduct a set of experiments and compare its performance to existing benchmark approaches. We also provide theoretical results comparing classical k-means with \texttt{DWF k-means}.  In particular, we show how \texttt{DWF k-means} generalises (distributed) classical k-means and how the weights follow naturally from this generalisation. We further discuss and address the problem of handling empty clusters and provide an example where this makes a difference.

\subsection{Related Work} 
Traditional clustering techniques have been extensively studied and employed to group similar data points into coherent clusters and uncover patterns within datasets. One of the simplest yet most popular of these is Lloyd’s method \cite{lloyd1982least} for k-means clustering, which is also at the core of our federated clustering method.
%K-means starts with a random initial set of k cluster centers, assigns each data point to the center nearest to it and recalculates the cluster centers to be the mean of all assigned points, iteratively until convergence. 
Other popular types of traditional clustering techniques include hierarchical clustering \cite{murtagh2011ward, murtagh2012algorithms}, density-based clustering \cite{ester1996density} and model-based clustering \cite{banfield1993model, fraley2002model}. However, these methods all assume that data is centralized, making them unsuitable for the increasingly prevalent scenarios where data privacy, security, and distribution are important.

It is important to distinguish between federated clustering and federated learning using clustering (also called clustered federated learning). Clustered federated learning focuses on employing clustering algorithms with the classical supervised FL workflow to categorise clients into clusters according to some similarity. This task has been studied in \cite{ghosh2019robust, sattler2020clustered, ghosh2020efficient, briggs2020federated, Kim2020DynamicCI, Xie2020MultiCenterFL}, among many others.
On the other hand, federated clustering refers specifically to methods that train clustering models in a FL environment.
K-means clustering was first extended to this type of federated clustering setting in \cite{kumar2020federated}, where the authors used an equally weighted mean of local cluster centers as an aggregation function to update global cluster centers. While the aggregation is the same as in our presented \texttt{Equally Weighted Federated k-means}, the training was slightly different.
Following this, \cite{pedrycz2021federated} and \cite{stallmann2022towards} introduced fuzzy c-means versions of the algorithm for the federated learning setting. Another fuzzy clustering algorithm called \texttt{federated fuzzy k-means clustering} was proposed in \cite{Wang2023FederatedFK}.
In \cite{dennis2021heterogeneity}, a single shot variant of federated k-means  called \texttt{k-FED} was proposed, focusing on communication-efficient clustering for heterogeneously distributed data. Their method requires each client to run classical k-means locally and communicate their local cluster means to a central instance, which then executes classical k-means to aggregate the results for a final clustering. Under certain, arguably realistic, assumptions of cluster separation and data distribution, \texttt{k-FED} can generate valid clusters in a single round of communication.
Based on a reformulation of the k-means problem into the matrix factorization (MF) model, which is an integer-constrained MF optimization problem, \cite{Wang2020FederatedCV} proposed an algorithm called \texttt{FedC}.
It relaxes the constraints by adding regularisation terms and run multiple steps of Projected Gradient Descent (PGD) locally and share them via Differential Gradient Sharing with the server.

On October 2023, Garst et al. \cite{Garst2023FederatedKC} published an article on arXiv proposing an algorithm called \texttt{Federated k-means clustering}. It uses similar weights our proposed algorithm \texttt{DFW k-means} works with. However, their algorithm works very differently from ours, and they use the weights in a different manner. Instead of utilizing the weights in the aggregation, as our approach does, the server performs weighted k-means on the received local cluster centroids in their work. The solitary disparity to \texttt{k-FED} in this step is therefore the replacement of k-means by weighted k-means. Moreover, they have an interesting approach to tackle the problem of empty clusters during training by allowing $k'\leq k$ locally, which does not work in our method and is hence different from our approach.
Despite the similarity of the weights, both works were developed independently, as our research was completed prior to the publication of theirs.

An algorithm for differential private vertical federated k-means based on the work of \cite{Ding2016KMeansCW} was proposed in \cite{Li2022DifferentiallyPV}. In our work, we focus on the horizontal setting.
Aside from the approaches based on classical clustering algorithms, there have also been recent advances using model-based methods. Chung et al. proposed an algorithm called \texttt{Federated Unsupervised clustering with Generative Models (UIFCA)}, a generative model extending the approach of \cite{ghosh2020efficient} to the unsupervised and heterogeneous federated learning setting.

There is also the related area of parallel and distributed clustering \cite{dhillon2002data, xu2002fast, tasoulis2004unsupervised}. 
Distributed clustering typically has higher communication costs, assumes that the data is IID distributed or gets distributed to workers in an IID way and is not driven by privacy in the first place. Consequently, parallel and distributed clustering, and federated clustering have distinct areas of application. Our research concentrates on the federated clustering setting.

\subsection{Structure of the Paper}
Following this introduction, section \ref{sec:chapter_flkm} details the \texttt{DWF k-means} algorithm and the theory behind our proposed approach. Section \ref{sec:chapter_experiments} describes the experimental setup and results for the various empirical studies that were done. Datasets, methodology, and hardware are detailed, followed by an analysis and discussion of results from the experiments. Finally, section \ref{sec:chapter_conclusion} concludes with a summary and suggestions for future work.
%%%%%%%%%%%%%%%%%%%%%%%%%%%%%%%%%%%%%%%%%%%%%%%%%%%%%%%%%%%%%%%%%%%%%%%%%%%%%%%%%%%%
%\section{Materials and Methods}
\section{The Dynamically Weighted Federated k-Means Algorithm}
\label{sec:chapter_flkm}
% ---------------------------------------------------
% ---------------------------------------------------
\subsection{Overview and Motivation}
K-means \cite{lloyd1982least} is a widely used unsupervised machine learning algorithm that aims to partition a dataset into a predetermined number of clusters. The algorithm iteratively refines cluster centroids to minimize the sum of squared distances between data points and their respective cluster centroids. K-means has found applications in various domains, including face detection, image compression and shape recognition \cite{su2001modified, manju2018ac, bai2018ensemble}.

Adapting the concept of \texttt{Federated Averaging} \cite{mcmahan2017communication} to the setting of k-means presents a challenge, primarily stemming from the disparity between the two algorithms' characteristics. \texttt{Federated Averaging} has traditionally been applied to neural networks, a domain characterized by gradient-based optimization. In the context of k-means, however, the challenge is to reconcile the differences introduced by the presence of cluster centers, a core element of the algorithm. Unlike neural networks, where model weights are updated through gradients, k-means involves updating cluster centroids based on data assignments.

\texttt{Federated Averaging} was initially motivated as an extension, in some sense, of the concept of large batch synchronous stochastic gradient descent referred to as \texttt{FederatedSGD} \cite{mcmahan2017communication, chen2016revisiting}. In synchronous SGD, a central server orchestrates model updates by aggregating gradients computed locally by participating devices. This synchronization ensures that all devices update their models concurrently, leading to more rapid convergence.

In the context of \texttt{Federated Averaging}, the traditional synchronous SGD approach is adapted to the decentralized setting of federated learning. Instead of a single round of local computation, \texttt{Federated Averaging} allows for multiple local rounds of computation on each device. This adjustment serves to capitalize on the computational capacity of each device while also accounting for the inherent variability in device capabilities and network conditions.

Our approach follows a similar trajectory by extending the concept of distributed updates to the centroids across the client devices. In this manner, each client independently performs a classical k-means step locally. 
Under certain assumptions, this decentralized algorithm is mathematically analogous to performing a single classical k-means step on the union of the clients' datasets.

The main difference between \texttt{Federated Averaging} and our approach is in the choice of weights. While \texttt{Federated Averaging} chooses time-independent weights proportional to the size of the local dataset, if that information is available to the server, we see (in section \ref{sec:results and discussion}) that carefully chosen dynamically computed weights improve convergence and performance of the federated version of k-means. For this reason, we term our approach \texttt{Dynamically Weighted Federated k-means} (\texttt{DWF k-means}).

\subsection{Distributed k-Means}
We first motivate how the steps performed by classical k-means can be distributed to multiple workers without changing the result, thus obtaining a distributed version of k-means. We denote vectors by lowercase bold letters $\mathbf{v} \in \R^m$ and matrices by uppercase bold letters $\mathbf{M} \in \R^{n \times m}$.
Let  $\textbf{c}_{j}^{(t)} \in \mathbb{R}^{m}$ be the global centroid number $j \in [k]:= \{1,...,k\}$ at time $t$ occurring in the computation of classical k-means and $X \subseteq \R^{m}$ be a sufficiently large dataset.
For simplicity, we define 
\begin{align*}
    X_{j}^{(t+1)} &:= \{\mathbf{x}\in X|\  \mathbf{x}\text{ is assigned to } \mathbf{c}_{j}^{(t)}\}  \\
    \omega_j^{(t)} &:= |X_j^{(t)}|.
\end{align*}
By "assigned" we mean that the centroid is the closest of all $k$-many centroids to a given data point $\mathbf{x}$ in this step, i.e.,
\begin{equation*}
    \mathbf{x} \text{ is assigned to } \textbf{c}_{j}^{(t)} :\iff j = \arg\min_{l \in [k]} \|\textbf{c}_{l}^{(t)} - \mathbf{x}\|_2.
\end{equation*}
In other words, $X_{j}^{(t)}$ is by definition the cluster $j$ at time $t$.
With these definitions, a classical k-means update of the centroid $\textbf{c}_j$ can be expressed as
\begin{equation}
\label{eq:flkm-one_classical_k_means_step}
    \textbf{c}_j^{(t)} = \frac{1}{\omega_j^{(t)}}\sum_{x \in X_{j}^{(t)}} x.
\end{equation}
Note that we assumed that the cluster $X_j^{(t)}$ is not empty, a case which needs to be treated carefully, especially in the federated setup as we will see.

Our goal is to reformulate this classical k-means step in a distributed manner. We assume that our data $X = \cup_{i=1}^N X_i$ is distributed to the clients with indices $i \in [N] = \{1,...,N\}$, where $X_i \subseteq \R^{m}$ is the local data of client $i$. We extend the definitions of $X_{j}^{(t+1)}$ and $\omega_j^{(t)}$ to
\begin{align*}
    X_{i, j}^{(t+1)} &:= \{\mathbf{x}\in X_i|\  \mathbf{x}\text{ is assigned to } \mathbf{c}_{j}^{(t)}\} \\
    \omega_{i, j}^{(t)} &:= |X_{i, j}^{(t)}|.
\end{align*}
If we assume that the $X_i$ are pairwise disjoint, the following equation holds
\begin{equation*}
    \omega_{j}^{(t)} = \sum_{i=1}^N \omega_{i,j}^{(t)}.
\end{equation*}
Under this assumption Equation \ref{eq:flkm-one_classical_k_means_step} can be rewritten as
\begin{align}
\label{eq:flkm_distributed_k_means_same_as_classical_k_means}
    \textbf{c}_j^{(t)} &= \frac{1}{\omega_{j}^{(t)}} \sum_{\mathbf{x} \in X_{j}^{(t)}} \mathbf{x} \nonumber \\
    &= \frac{1}{\omega_{j}^{(t)}} \sum_{i=1}^N \omega_{i, j}^{(t)} \left(\frac{1}{\omega_{i, j}^{(t)}}\sum_{\textbf{x} \in X_{i, j}^{(t)}} \textbf{x}\right) \\
    & = \sum_{i=1}^N \lambda_{i, j}^{(t)} \cdot \textbf{c}_{i, j}^{(t)},  \nonumber 
\end{align}
where $\textbf{c}_{i, j}^{(t)}:=\frac{1}{\omega_{i, j}^{(t)}}\sum_{\textbf{x} \in X_{i, j}^{(t)}} \textbf{x}$ is by definition the (local) centroid obtained by performing one classical k-means step on the local dataset $X_i$ of all clients, starting with the previous (global) centroids $\textbf{c}_{1}^{(t-1)},...,\textbf{c}_{k}^{(t-1)}$ and $\lambda_{i, j}^{(t)} := \frac{\omega_{i, j}^{(t)}}{\omega_{j}^{(t)}}$.

We see that one global step of classical k-means on the full dataset $X$ leads to exactly the same centroids as performing one classical k-means step on all clients and aggregating these local centroids via a weighted average with weights $\lambda_{i, j}^{(t)}$.

\subsection{From Distributed to Federated k-Means}
As mentioned at the beginning of the chapter, Equation \ref{eq:flkm_distributed_k_means_same_as_classical_k_means} can be used to generalize k-means to a federated setting, similar to the generalization from \texttt{FederatedSGD} to \texttt{Federated Averaging}, by allowing multiple classical k-means steps on each client and relaxing the assumption that each client participates. 

Let $\textbf{c}_{i, j}^{(t)} \in \mathbb{R}^{m}$ denote the local centroid of index $j \in [k]$ at global time $t$ locally computed on client $i \in [N]$ and $\omega_{i, j}^{(t)} \in \mathbb{R}_{\geq 0}$ be some weights, both of which will be defined fully later. We define the aggregated global centroid $\textbf{c}_j^{(t)} \in \mathbb{R}^{m}$ by
\begin{align}
\label{eq:flkm_definition_aggregation_dwfkM}
    \textbf{c}^{(t)}_j := \sum_{i \in I_t} \lambda_{i, j}^{(t)} \cdot \textbf{c}_{i, j}^{(t)},
\end{align}
with
\begin{align}
    \lambda_{i, j}^{(t)} := 
    \begin{cases}
        \frac{ \omega_{i, j}^{(t)}}{\sum_{i \in I_t} \omega_{i, j}^{(t)}}, & \text{if } \sum_{i \in I_t} \omega_{i, j}^{(t)} > 0 \\
        \frac{1}{|I_t|}, & \text{else,}
    \end{cases} 
\end{align}
where $\emptyset \neq I_t \subseteq \{1,...,N\}$ is the set of indices of participating clients at global round $t>0$. The definition of $\lambda_{i, j}^{(t)}$ handles the case where all the weights are zero, in which we take the equally weighted average over all participating clients as a fallback. This fallback is important because this case can occur more frequently when the number of clients participating per round is relatively small and the data is distributed in a non-IID manner. How to deal with centroids when the associated cluster is empty and, therefore, the associated weight is zero is discussed in the next section.

If only a small randomly selected non-empty subset $I_t \subseteq [N]$ of clients participate at iteration $t$, which is the case in many federated learning settings, this approach could typically result in relatively high noise on the $\textbf{c}_j^{(t)} \in \mathbb{R}^{m}$. This is especially true in cases where the data is distributed very heterogeneously.
To dampen this effect, we modify Equation \ref{eq:flkm_definition_aggregation_dwfkM} by introducing a learning rate $\eta \in (0, 1]$ and a momentum $\mu \in [0, 1)$. Let 
\begin{align}
    \textbf{d}^{(t)}_j := \sum_{i \in I} \lambda_{i, j}^{(t)} \cdot \textbf{c}_{i, j}^{(t)}
\end{align}
be the aggregated centroids as defined previously.
The update of the global centroids, including learning rate and momentum, is defined by
\begin{align}
    \textbf{c}^{(t+1)}_j = \textbf{c}^{(t)}_j + \eta \cdot  \left(\textbf{d}^{(t + 1)}_j -  \textbf{c}^{(t)}_j\right)+ \mu \cdot \left(\textbf{c}^{(t)}_j-\textbf{c}^{(t-1)}_j\right),
\end{align}
with the convention $\textbf{c}^{(-1)}_j := \textbf{c}^{(0)}_j$.

A crucial part of this algorithm is the choice of the weights. The weights must take into account that
\begin{itemize}
    \item the amount of data on each client could be different (non-IID in the size of local data),
    \item the distribution of the data itself could be different on each client (non-IID in the distribution).
\end{itemize}
As discussed in the motivation, a suitable choice for the weights, at least if only a single classical k-means step is performed locally on each client, is given by
\begin{align}
\label{eq:flkm_definition_weights}
    \omega_{i, j}^{(t+1)} := |\{\textbf{x} \in X_i|\ \textbf{x} \text{ is assigned to centroid } \textbf{c}_{j}^{(t)}\}|.
\end{align}
These weights attempt to compensate for the imbalance of the data and are, therefore, well-suited to the federated learning setting. However, a drawback of these weights is that they disclose certain details regarding the local data. This is discussed in more detail in the privacy section \ref{sec:chapter_privacy}. When using these weights, we term this algorithm \texttt{Dynamically Weighted Federated k-Means (DWF k-means)}.
Alternatively, one could weight each centroid equally, i.e., 
\begin{align}
\label{eq:flkm_definition_weights_equally}
    \omega_{i, j}^{(t)} := \frac{1}{|I_t|}.
\end{align}
When these weights are used in each step, we call the algorithm \texttt{Equally Weighted Federated k-Means (EWF k-means)}. It is comparable to the \texttt{Federated Averaging} algorithm. 
Completely different weights are also conceivable, and in the case where we leave the choice of weights open, we simply refer to it as \texttt{Weighted Federated k-means (WF k-means)}.

The remaining algorithm works exactly as one would expect from a classic k-means algorithm in a centralized federated learning setting.
Global centroids $\textbf{c}^{(0)}_j\in \mathbb{R}^{m}$ for all $j=[k]$ are initialized and communicated to the clients, where a small number of classical k-means steps are performed. This results in a set of centroids $\textbf{c}_{i, j}^{(t)} \in \mathbb{R}^{m}$ for each participating client $i \in I_t$. These local centroids are sent back to the server and aggregated to obtain the new global centroids. This process is referred to as a single \textit{global learning round}. The steps of the algorithm where the clients train their local model during one global training round are called \textit{local learning rounds}. This process is repeated until the algorithm has converged, which is the case if the movement $\|\mathbf{C}^{(t)}-\mathbf{C}^{(t-1)}\|_F < \varepsilon$ is under a given threshold $\varepsilon>0$. Here, $\|\cdot\|_F$ denotes the Frobenius norm and $\mathbf{C}^{(t)}, \mathbf{C}^{(t-1)}$ are the matrices whose columns consists of the centroids $\textbf{c}^{(t)}_1,...,\textbf{c}^{(t)}_k$ and  $\textbf{c}^{(t-1)}_1,...,\textbf{c}^{(t-1)}_k$ respectively.

The number of required iterations to convergence depends crucially on the choice of initial centroids. We suggest using \texttt{k-FED} \cite{dennis2021heterogeneity} to obtain an initial set of centroids. The advantage of \texttt{k-FED} is that it is a single-shot algorithm and, therefore, adds only a small overhead to the \texttt{WF k-means} algorithm. The centroids calculated by \texttt{k-FED} are typically relatively close to the optimal centroids, drastically reducing the number of iterations to convergence.

The main algorithm is outlined in Algorithm \ref{alg:dwfkm}, while the required sub-algorithms can be found in the Appendix \ref{appendix:algorithms}.
% ---------------------------------------------------
\begin{algorithm}[h]
\caption{Weighted Federated k-Means}
\label{alg:dwfkm}
\SetKwInOut{Input}{Input}\SetKwInOut{Output}{Output}
\SetKwProg{ForAllInParallel}{foreach}{ do in parallel}{end} 
\Input{
 Local datasets $(X_{i})_{i \in [n]}$, 
 number of clusters $k$, number of runs $n_{\text{init}}$, 
 maximal number of global iterations per run $\text{max}_{\text{global}}$,  
 maximal number of local iterations per global round $\text{max}_{\text{local}}$,
 learning rate $\eta$, 
 momentum $\mu$, 
 tolerance $\epsilon >0$, 
 number of participating clients per round $n_{\text{clients}}$.}
\Output{Centroids as $m \times k$ matrix, where $m$ is the dimension of the data.}
best\_centroids, best\_score $\leftarrow$ None, $\infty$\;
\For{$\text{iteration} = 1,...,n_{\text{init}}$}{
$\textbf{C} \leftarrow$ InitialCentroids()\;
$\textbf{C}_{\text{old}} \leftarrow$ \textbf{C}\;
\For{$t =$ 1,..., $\text{max}_{\text{global}}$}{
    $I_t \leftarrow$ ChooseRandomClients($n_{\text{clients}}$)\;
    \ForAllInParallel{i $\in I_t$}{
        $\textbf{C}_{i}$, $\omega_{i}$ $\leftarrow$ LocalUpdate($X_{i}$, $\textbf{C}$, $\text{max}_{\text{local}}$, $\epsilon$)\;
        }
     $\textbf{C}_{\text{new}} \leftarrow$ Aggregte($(\textbf{C}_{i})_{i \in I_t}$, $(\omega_{i})_{i \in I_t}$, \textbf{C}, $\textbf{C}_{\text{old}}$, $\eta$, $\mu$)\;
    $\text{movement} \leftarrow \norm{\textbf{C}_{\text{new}} - \textbf{C}}_F$ \;
    $\textbf{C}, \textbf{C}_{\text{old}}  \leftarrow \textbf{C}_{\text{new}}, \textbf{C}$\;
    \uIf{$t \geq \text{max}_{\text{global}}$ or $\text{movement} < \epsilon$}{
     score $\leftarrow$ RequestScore($(X_{i})_{i \in [n]}$, $\textbf{C} $)\;
     \uIf{score $<$ best\_score}{
        best\_score, best\_centroids $\leftarrow$ score, $\textbf{C}$\;
     }
    \textbf{break}\;
    }
}
}
\textbf{return} best\_centroids\;
\end{algorithm}

%%%%%%%%%%%%%%%%%%%%%%%%%%%%%%%%%%%%%%%%%%%%%%%%%%%%%%%%%%%%%%%%%%%%%%
\subsection{Handling empty clusters}
\begin{figure}[h]
    \centering
    \includegraphics[width = 0.9\textwidth, trim={2.5cm 1.5cm 2cm 1.5cm},clip]{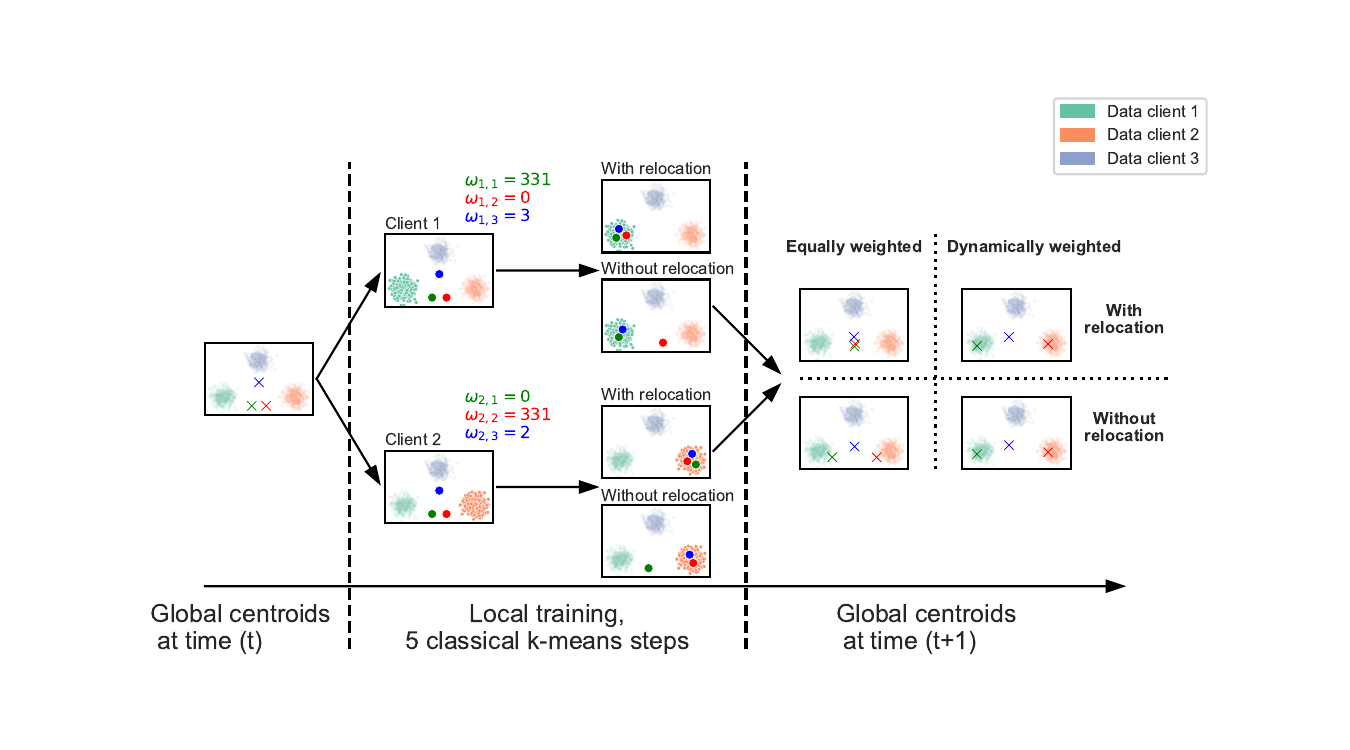} 
    \caption{The illustration depicts one round of global learning in which 2 of the 3 clients participate in the round at time $t$. Locally, 5 classical k-means steps are performed. The data is distributed in a non-IID manner, implying that each data blob is associated with a different client. Four scenarios are presented, comparing local training with and without relocating centroids for empty clusters, and aggregating equally weighted and dynamically weighted approaches. It should be noted that the weights used are based on the latest global centroids and can be computed directly upon receipt by clients. The weights only apply to aggregation in the dynamically weighted model and do not affect local training.}
    \label{fig:flkm-weights-relocating}
\end{figure}

Many implementations of classical k-means, such as that of Scikit-Learn \cite{scikit-learn}, tackle the problem of empty clusters while updating the centroids by relocating the centroid to some more or less random data point, if no data point is assigned to that centroid at some stage of the algorithm. While this is beneficial in the classical case, this behavior could have a negative influence on the quality of the outcome of the \texttt{WF k-means} algorithm or the convergence rate. If, for instance, not all clients participate in every round, which is typical in federated learning, and the data is non-IID in the distribution, it could happen somewhat often that some centroids have no data points assigned to them. However, it is possible that a specific centroid aligns optimally with the union of the local datasets, but the clients that possess the relevant data are not participating in a particular global training round. Therefore, relocation would move the well-placed centroid due only to the fact that the relevant clients are not currently participating. While this effect plays a minor role if the adapted weights from \texttt{DWF k-means} are used for aggregation, it is important if the equally weighted version of the aggregation is used, which is also the fallback of the \texttt{DWF k-means} algorithm. These considerations are illustrated with a small toy example in Figure \ref{fig:flkm-weights-relocating}.
%-----------------------------------------------------------------------------
%-----------------------------------------------------------------------------
%%%%%%%%%%%%%%%%%%%%%%%%%%%%%%%%%%%%%%%%%%%%%%%%%%%%%%%%%%%%%%%%%%%%%%%%%%%%%%%%%%
\subsection{Privacy} 
\label{sec:chapter_privacy}
Federated learning emerged as a direct response to privacy concerns associated with traditional centralized machine learning paradigms.
In recent years, data privacy has become increasingly important, not least due to legal regulations such as the GDPR. 
Federated learning seeks to bridge the gap between collaborative data analysis and preserving individual data privacy. Today it is used in a number of customer products and services like Google's Gboard \cite{yang2018applied, chen2019federated}, Apple's vocal classifier for "Hey Siri" \cite{Apple2019designforprivacy} and many others \cite{leroy2019federated, EUCORDISDrug, erlingsson2014rappor}.

While federated learning is an innovative approach to decentralized model training, it is important to note that it does not provide formal proof to ensure privacy. This is not surprising since the global model can mostly be examined as in the classic centralized case and be subject to various attack scenarios. Classical threat models like membership inference, reconstruction, property inference, and model extraction attacks \cite{al2019privacy, hu2022membership, rigaki2020survey} can also be applied to the federated learning setting. Furthermore, new attack possibilities emerge in this setting, such as a curious or malicious server or malicious clients, leading to privacy leakage \cite{wang2019beyond, mothukuri2021survey} and other security issues \cite{bagdasaryan2020backdoor, sun2019can, wang2020attack}.
Therefore, to meet privacy requirements, other protection mechanisms are nowadays combined with federated learning. These techniques include Differential Privacy \cite{dwork2006calibrating, dwork2008differential, dwork2014algorithmic, wei2020federated}, Secure Multi-Party Computation \cite{yao1986generate, mohassel2017secureml}, Homomorphic Encryption \cite{gentry2009fully, zhang2020batchcrypt}, and Secure Aggregation \cite{bonawitz2017practical, bell2020secure}.

However, the previously mentioned topics mostly apply to neural networks. In our case, the analogous question is how much information the dynamically calculated weights and centroids sent to the central server reveal about the local datasets $X_i$ of the clients. The dynamic weights determine exactly how many data points are assigned to the centroids that the server sends to the clients. A curious server could, therefore, collect all the weights over time and use this information to reconstruct parts of the local distribution of the client's data. In addition, a malicious server could send selected centroids to the clients with the intention of learning as much as possible about the local data from the returned information.

It was not within the scope of this work to quantify the leakage of information arising from the sharing of the centroids and dynamic weights, or to investigate whether one could recover at least parts of the local data $X_i$ by collecting the centroids and weights over time. These are tasks left for future work. We merely outline a few concepts here, that could potentially enhance the privacy of \texttt{WF k-means}, without providing extensive details. 

One possible approach would be to limit the number of times a client can participate in the training. This reduces the amount of data transmitted to the server which, in turn, limits the amount of information that the server can learn about the local data. Another option would be to round each weight to a specific number of decimal places rather than sending back the exact weights. However, this approach may offer only a minor improvement in terms of privacy. Similarly, one can utilize \texttt{EWF k-means} as an alternative to \texttt{DWF k-means} for minimizing the transmitted information to the server.
 Moreover, one could add noise to the centroids and/or the weights, as is usually done to guarantee differential privacy for some $\varepsilon>0$. Finally, schemes such as secure aggregation or homomorphic encryption could be adopted in such a way that the centroids and weights are sent encrypted or masked to the server, so that it does not receive any valuable information about them.
%%%%%%%%%%%%%%%%%%%%%%%%%%%%%%%%%%%%%%%%%%%%%%%%%%%%%%%%%%%%%%%%%%%%%%%%%%%%%%%%%%%%
%%%%%%%%%%%%%%%%%%%%%%%%%%%%%%%%%%%%%%%%%%%%%%%%%%%%%%%%%%%%%%%%%%%%%%%%%%%%%%%%%%%%
%%%%%%%%%%%%%%%%%%%%%%%%%%%%%%%%%%%%%%%%%%%%%%%%%%%%%%%%%%%%%%%%%%%%%%%%%%%%%%%%%%%%
\section{Materials and Methods} 
\label{sec:chapter_experiments}

%%%%%%%%%%%%%%%%%%%%%%%%%%%%%%%%%%%%%%%%%%%%%%%%%%%%%%%%%%%%%%%%%%%%%%%%%%%%%%%%%%%%
\subsection{Algorithms}
For comparison, we trained four algorithms using varying datasets and distributions. Firstly, we trained classical k-means as a baseline on the complete dataset, which is the union of the clients' datasets. Secondly, we trained two weighted federated k-means models on the considered datasets, one with equal weights (\texttt{EWF k-means}) and one with dynamic weights (\texttt{DWF k-means}).
To compare our algorithm with other clustering algorithms in the federated learning setting, we chose \texttt{k-FED} as the fourth algorithm due to its simplicity of implementation. 
However, as not all clients should participate in the training, we made slight adjustments to \texttt{k-FED} to ensure that only the selected portion of clients participate during training. In the federated learning setting, models are typically trained iteratively over multiple rounds. \texttt{k-FED}, on the other hand, is a single-shot algorithm. While it is common for only a small subset of clients to participate in each global training round in the iterative case, \texttt{k-FED} typically involves more clients in training due to its single-shot nature. This makes it difficult to compare results if not all clients are used in the training. However, since the chosen proportions of participating clients will vary from case to case and will strongly depend on the use case, we also trained \texttt{k-FED} with the same number of participating clients as the two \texttt{WF k-means} algorithms. If all clients participate in each training round, the results are directly comparable.

%%%%%%%%%%%%%%%%%%%%%%%%%%%%%%%%%%%%%%%%%%%%%%%%%%%%%%%%%%%%%%%%%%%%%%%%%%%%%%%%%%%%
\subsection{Datasets and Distributions}
\begin{figure}[h]
    \centering
    \begin{subfigure}{.33\textwidth}
    \captionsetup{justification=centering}
      \centering
      \includegraphics[width = 0.9\textwidth, trim={0 1cm 0 1cm}]{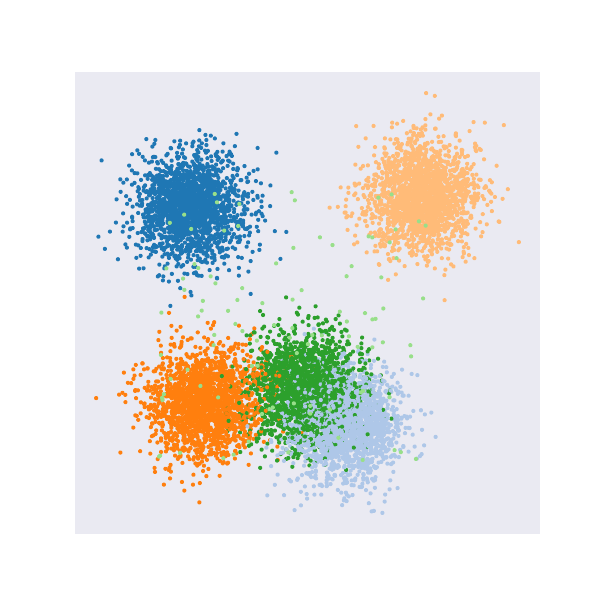}
      \caption{Synthetic data}
    \end{subfigure}%
    \begin{subfigure}{.33\textwidth}
    \captionsetup{justification=centering}
      \centering
      \includegraphics[width = 0.9\textwidth, trim={0 1cm 0 1cm}]{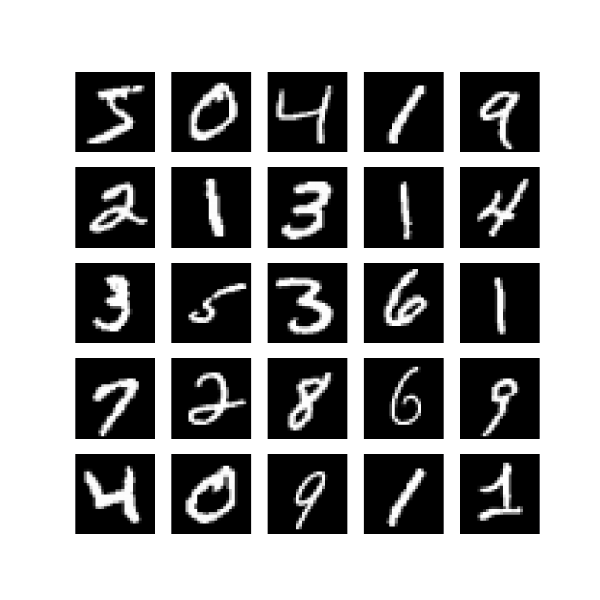}
      \caption{MNIST}
    \end{subfigure}%
    \begin{subfigure}{.33\textwidth}
    \captionsetup{justification=centering}
      \centering
      \includegraphics[width = 0.9\textwidth, trim={0 1cm 0 1cm}]{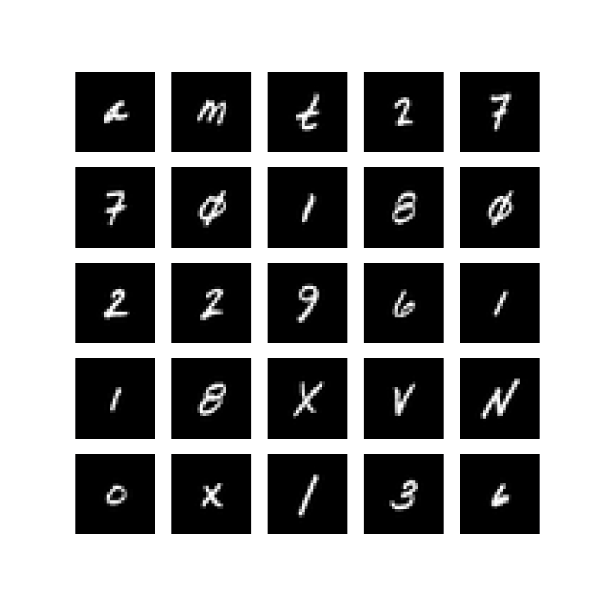}
      \caption{FEMNIST}
    \end{subfigure}%
\caption{The synthetic data in (a) was created with Scikit-learn. The dataset contains 10,100 points, including 100 random data points. The color of each data point corresponds to its true label. 25 random images from the MNIST and FEMNIST datasets are shown in Figures (b) and (c). All images within the MNIST and FEMNIST datasets are greyscale and consist of $28 \times 28$ pixels.}
\label{fig:datasets}
\end{figure}

We trained the algorithms on three datasets suitable for classification, a common task where clustering is used: a simple synthetically generated dataset, MNIST \cite{lecun2010mnist}, and Federated Extended MNIST (FEMNIST) \cite{caldas2018leaf}, which is a modification of the MNIST dataset specially developed to test federated learning algorithms. Figure \ref{fig:datasets} provides an overview of the datasets used.

The synthetic dataset consists of five two-dimensional blobs with 10,000 samples in total and is created with Scikit-learn \cite{scikit-learn}. The centers $\mathbf{c}$ are chosen randomly in $\mathbf{c} \in [-1,1]^2$ from a uniform distribution in each component and the standard deviation of the isotropic Gaussian blobs is $0.2$. To test stability, we added 100 uniformly distributed, randomly selected data points $\mathbf{x} \in [-1,1]^2$. The synthetic data was chosen because it is simple enough to interpret the behavior of different algorithms and to visualize the results and steps during development.

The second dataset we considered is MNIST, a standard dataset in the field of machine learning and computer vision. This dataset consists of a diverse collection of $28 \times 28$ pixel greyscale images, each representing handwritten digits ranging from 0 to 9. The dataset consists of 60,000 training samples and 10,000 test samples, however, we only used the training data for training and evaluation. 

In the case of the synthetic dataset and MNIST, we distributed the data to 100 clients in three different ways: IID, half-IID, and non-IID. For the IID case, we randomly reordered the data and then split the data into 100 chunks of equal size. Each chunk was treated as the data of a specific client. All clients thus had the same distribution of data, both in terms of content and number. In order to test the algorithms in an extreme non-IID scenario, we distributed the data in the following way: We ran classic k-means on the dataset with k=100 with a maximum of 5 steps and $n_{\text{init}}=5$. Each cluster was then treated as the data of one client. The distribution of the data was therefore very heterogeneous, both in terms of content and amount of data. For the half-IID distribution, half of the data was distributed as in the IID case and the other half as in the non-IID case.

We also ran the experiments for these two datasets where we distributed the data to only 10 instead of 100 clients, the results of which can be found in Appendix \ref{appendix:results}.

\begin{figure}[h]
    \centering
    \includegraphics[width = 0.95\textwidth]{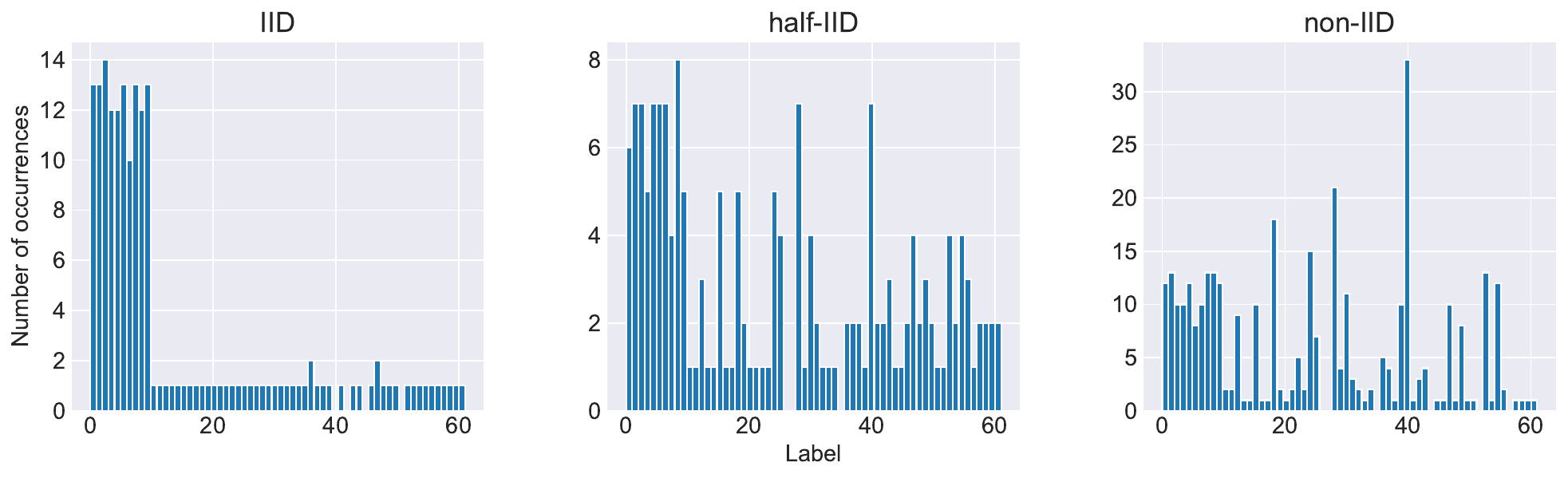}
    \caption{Histogram of the generated FEMNIST data for client 1 for each distribution: IID, half-IID and non-IID. The number of frequencies of the given 62 labels (10 digits, 26 lower case letters, and 26 upper case letters) is plotted on the y-axis.}
    \label{fig:distribution femnist client 1}
\end{figure}

The third dataset we used is FEMNIST, from the benchmark framework for federated settings called LEAF \cite{caldas2018leaf}. 
Analogous to MNIST, each image consists of $28 \times 28$ greyscale pixels. The dataset contains 62 different classes consisting of 10 digits, 26 lower case letters and 26 upper case letters. Unlike the other two datasets, FEMNIST is not a single, ready-to-use dataset. Instead, the data is generated using the LEAF framework according to the given parameters and is already distributed to the clients with the specified distribution. It provides the ability to create IID and non-IID data. We constructed the half-IID case by combining half of the IID data with an equal amount of data from the non-IID case for each client. We then took the first 100 clients for each distribution.
As parameters we used $\textit{fraction of data to sample} = 0.05$, $\textit{t} = \textit{"sample"}$, $\textit{sample seed} = 549786595$, and  $\textit{split seed} = 154978679$. Our FEMNIST IID dataset finally contained 16,744 samples (distributed over 100 clients), the half-IID dataset contained 16,434 samples and the non-IID dataset 28,585 samples.
As an example, the distribution of the first client's data for all three cases is shown in Figure \ref{fig:distribution femnist client 1}.
%-----------------------------------------------------------------------------
%-----------------------------------------------------------------------------
Note that since the distributions were generated differently for the synthetic dataset and MNIST, the results are not directly comparable to those from the FEMNIST experiments.
%%%%%%%%%%%%%%%%%%%%%%%%%%%%%%%%%%%%%%%%%%%%%%%%%%%%%%%%%%%%%%%%%%%%%%%%%%%%%%%%%%%%
\subsection{Metrics}
The k-means algorithm minimizes the objective function
\begin{align*}
    \text{score}(X, \mathbf{C}):= \frac{1}{|X|}\sum_{j=1}^k \sum_{\mathbf{x} \in X_j} \| \mathbf{x} - \mathbf{c}_j\|_2^2,
\end{align*}
where $X$ denotes the dataset, $\mathbf{C}$ the matrix whose columns are the centroids $\mathbf{c}_1,...,\mathbf{c}_k$ and $X_j$ the set of data points assigned to $\mathbf{c}_j$. Note that we scaled the score by a factor of $\frac{1}{|X|}$ to be independent of the size of the dataset $X$. Since k-means tries to optimize this score, it is one of the metrics we used to evaluate the algorithms.

We also used accuracy as a metric for the evaluation, as we have the ground truth of the labels. Accuracy is defined as the ratio of correct predictions in relation to the total number of cases under consideration. Since the clustering labels and the true labels do not have to match, we needed to find the correct assignment of these labels. To do this, we counted the occurrences of true labels $l_{\text{true}}$ in the samples corresponding to a given predicted label $l_{\text{pred}}$, and assigned the predicted label $l_{\text{pred}}$ to the true label $l_{\text{true}}$ that occurs most often. In other words, we determined the matching between true and predicted labels which maximizes the accuracy.

The external entropy-based cluster evaluation v-measure \cite{rosenberg2007vmeassure} is the third metric we used. It is defined as the harmonic mean of homogeneity and completeness:
\begin{align*}
    \text{v-measure} := 2 \cdot \frac{\text{homogeneity} \cdot \text{completeness}}{\text{homogeneity} + \text{completeness}},
\end{align*}
where homogeneity measures how much the class distribution within each cluster is skewed to a single class and completeness indicates how much a clustering assigns all of those data points that are members of a single class to a single cluster. For more details see \cite{rosenberg2007vmeassure}. The v-measure is hence a score between $0$ and $1$, where $1$ indicates perfect homogeneity and completeness. This metric is thus suitable for evaluating clustering.

%%%%%%%%%%%%%%%%%%%%%%%%%%%%%%%%%%%%%%%%%%%%%%%%%%%%%%%%%%%%%%%%%%%%%%%%%%%%%%%%%%%%
\subsection{Parameters}
\label{subsec:parameters}
For all three datasets, we distributed the data to 100 clients in three different ways: IID, half-IID and non-IID. The tolerance (globally and locally, if necessary) was set to $\epsilon = 10^{-8}$. For the experiments, we implemented a second stopping criterion in the \texttt{WF k-means} algorithms. The algorithms were assumed to have converged if the movement of the centroids did not decrease globally over the last $N_{\text{steps}} = 300$ steps. The number of maximal global and local training iterations was set to $\text{max}_{\text{global}}=10,000$, which was also used for classical k-means and \texttt{k-FED}, both globally and locally. The number of repetitions inside each algorithm was set to $n_{\text{init}}=1$ for all algorithms. For the two \texttt{WF k-means} algorithms, we set $\text{max}_{\text{local}}=5$. For the learning rate and momentum, we used $lr = 0.01$ and $m=0.8$. With these configurations, we ran the experiments with different numbers of participating clients per global training round (only for \texttt{EWF k-means}, \texttt{DWF k-means} and \texttt{k-FED}), namely $n_{\text{clients}} = 5,10,15,...,100$.
The number of clusters $k$ was chosen depending on the dataset. For the synthetic dataset, $k=5$ was used, for MNIST we used $k=20$, and for FEMNIST $k=64$.
The choice of $k$ was driven by the number of categories in the dataset and runtime of the experiments, not by heuristics such as the elbow method. This is appropriate since our aim is to compare the federated clustering algorithms with classical k-means clustering as a baseline, hence the concrete choice of $k$ plays a minor role. 

For each configuration, we trained each model 100 times. For the synthetic dataset and MNIST, the distribution of the data to the 100 clients was randomly generated for each of these 100 runs. For FEMNIST, the distribution to the clients was fixed as it was generated with LEAF. To compare the results and remove runs where the algorithms became stuck at local minima, we took the 50 best, and therefore lowest scoring runs and calculated the average and standard deviation for each metric on those runs.

In summary, given by the number of configurations used for each parameter, we trained in total
\[\text{datasets}\cdot\text{distributions}\cdot\text{models}\cdot\text{\#Clients per round}\cdot\text{runs}=3 \cdot 3 \cdot 4 \cdot 20 \cdot 100 = 72,000\]
models.
%%%%%%%%%%%%%%%%%%%%%%%%%%%%%%%%%%%%%%%%%%%%%%%%%%%%%%%%%%%%%%%%%%%%%%%%%%%%%%%%%%%%
\section{Results and Discussion}\label{sec:results and discussion}
We trained the models on four NVIDIA A100 processors, each having 24 CPU cores and 125 GB of RAM. The training duration for synthetic data was approximately one day, whereas the MNIST experiments took three days and the FEMNIST experiments took nineteen days to complete. For the implementation, we used Python 3.11.4, Scikit-learn 1.3.0, and Numpy 1.24.3.

\begin{figure}[h]
    \centering
    \includegraphics[width = 0.95\textwidth]{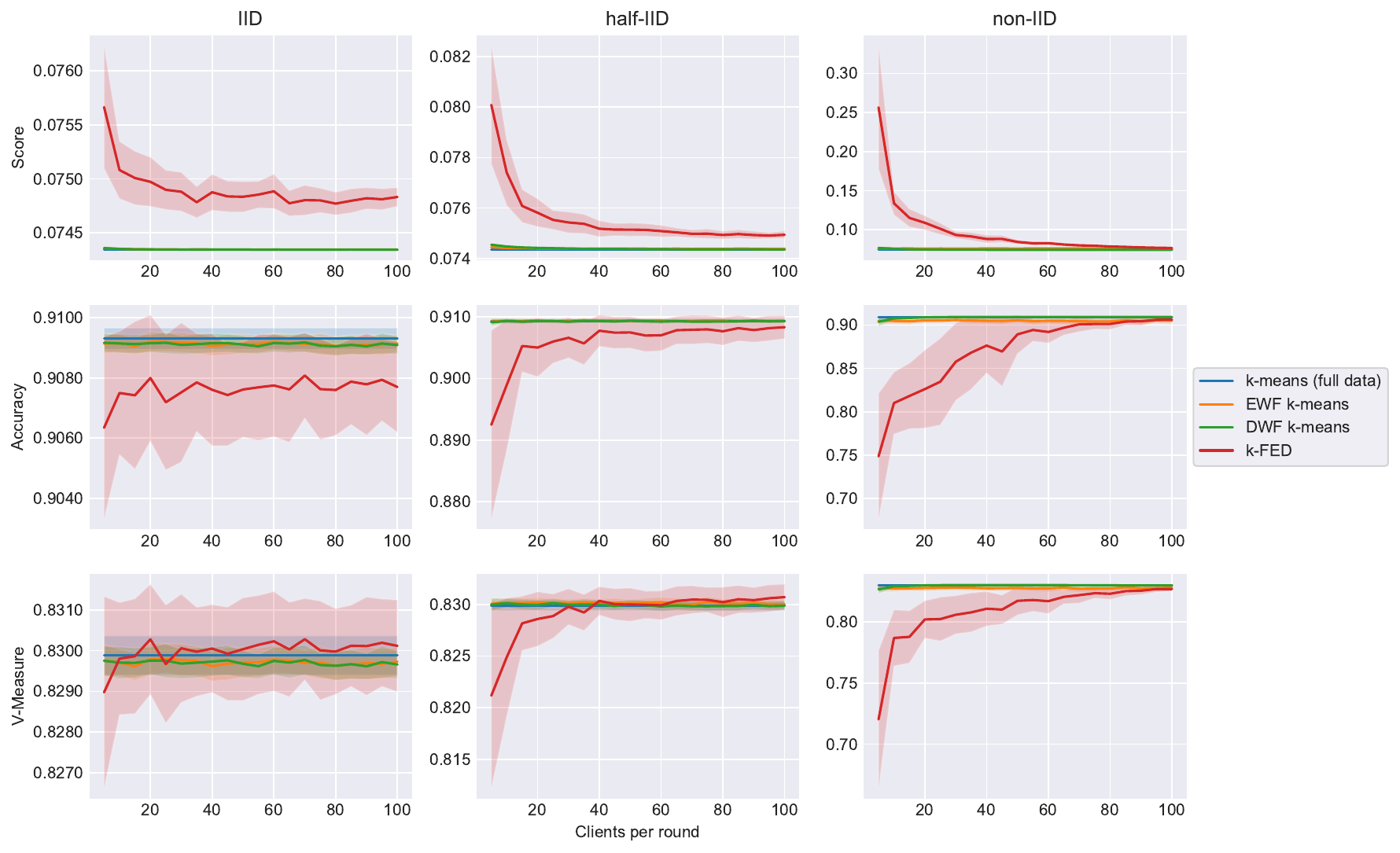}
    \caption{The results for the synthetic data. To remove results where the algorithms became stuck in local minima, for each of the values for $n_{\text{clients}}$ we only kept the 50 runs out of 100 with the lowest score. The solid line is the mean $m$ of the calculated metrics of the 50 runs (dependent on $n_{\text{clients}}$), the area around it is the mean $\pm$ the standard deviation of these 50 runs.}
    \label{fig:experiments synth 100 all metrics}
\end{figure}

\begin{figure}[h]
    \centering
    \includegraphics[width = 0.95\textwidth]{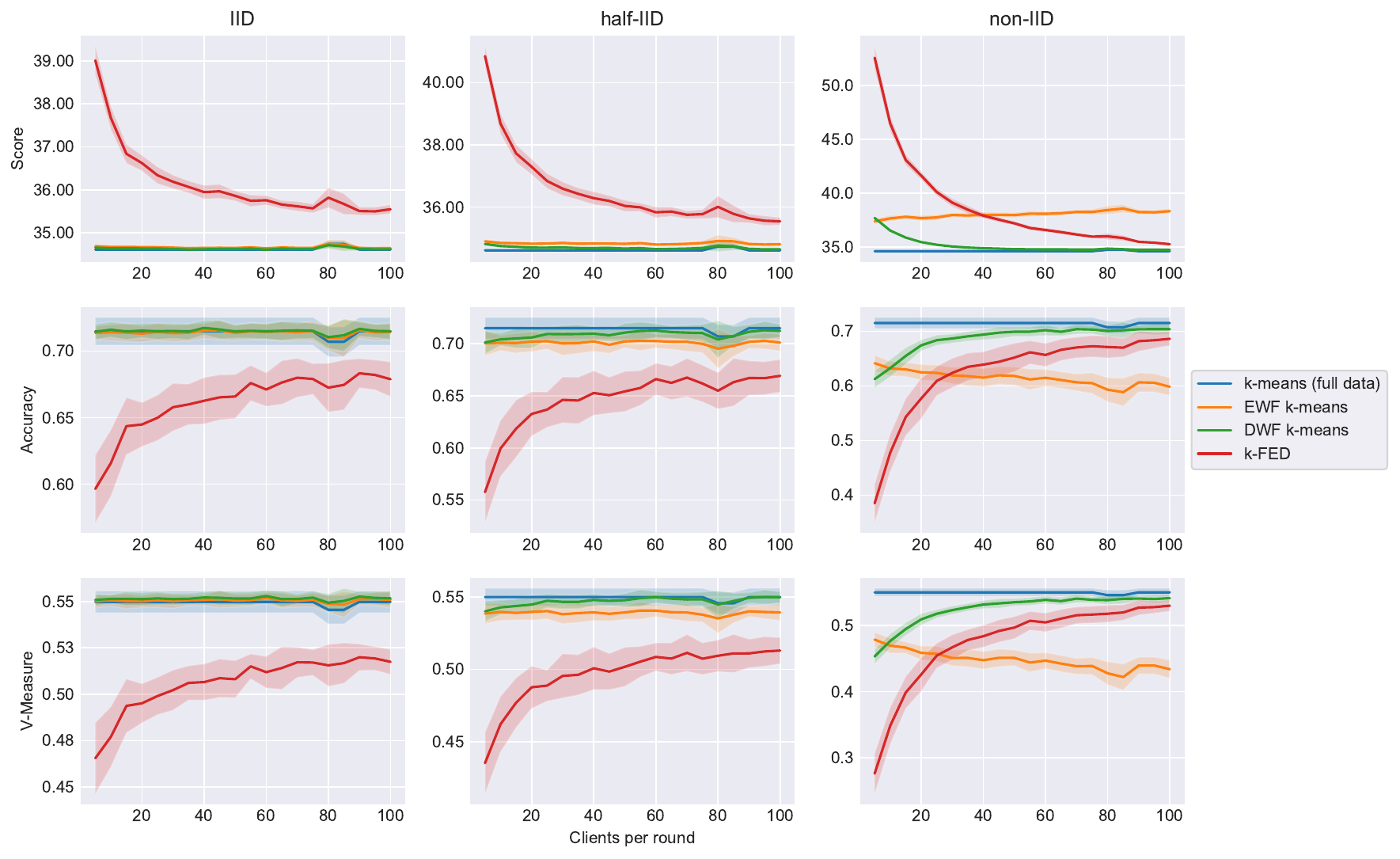}
    \caption{The results for MNIST. Analogue to Figure \ref{fig:experiments synth 100 all metrics}.}
    \label{fig:experiments MNIST 100 all metrics}
\end{figure}

The results for the synthetic data are shown in Figure \ref{fig:experiments synth 100 all metrics}.
Considering the classical k-means score and accuracy on the synthetic dataset, we observed that \texttt{EWF k-means} and \texttt{DWF k-means} are able to achieve similar results to classical k-means. As expected, they outperformed \texttt{k-FED} in the IID and half-IID settings, since \texttt{k-FED} is known to draw its advantages from heterogeneously distributed data \cite{dennis2021heterogeneity}. Therefore, we observed that the \texttt{k-FED} score comes close to those of classical k-means and \texttt{DWF k-means} only in the non-IID setting, when the percentage of participating clients reaches around 50 percent. Since \texttt{k-FED} is a one-shot method, focusing on communication efficiency, it has the disadvantage of seeing less data when the number of participating clients is small. However, the half-IID case is the most realistic of the three in many applications, since in reality complete homogeneity or an extreme imbalance as in the non-IID case are rather the exception. Hence, based on the score, \texttt{EWF k-means} and \texttt{DWF k-means} outperformed \texttt{k-FED} on the relatively easy synthetic dataset. 
Interestingly, in the case of v-measure, \texttt{k-FED} seemed to be on par or slightly outperformed the other methods in the IID setting, although this could be due to randomness since the scale we are considering here is very small and the variance in the scores is especially high for \texttt{k-FED}.
\FloatBarrier
More instructive are the results on the more complex MNIST dataset, as shown in Figure \ref{fig:experiments MNIST 100 all metrics}.
Here we observed a similar trend as for the synthetic data, where \texttt{EWF k-means} and \texttt{DWF k-means} achieved similar results to classical k-means, while \texttt{k-FED} performed worse than the rest in the IID setting. Interestingly, the \texttt{EWF k-means} performed significantly poorly on the non-IID data. This was due to the equal weighting causing centroids to be pulled equally in disparate directions, even where particular clients had only a small set of data points belonging to that particular centroid. So only in the non-IID case and only when at least 40 clients participated in each global training round, was \texttt{k-FED} able to outperform \texttt{EWF k-means}. However, it always lagged behind \texttt{DWF k-means}.

Also notable is the slight but consistently better performance of the \texttt{EWF k-means} when there are few clients participating per global training round in the non-IID setting. Due to the way the data is distributed across clients in the non-IID setting, the clients each have data mainly from a single cluster, with only a few data points belonging to other clusters. Hence, in the case of low client participation, where the client with the bulk of the cluster data is not present, equally weighting all participating clients could be helpful to prevent any artificial bias in the direction of a particular client's data distribution when that client still has a relatively small amount of data points belonging to a particularly underrepresented cluster.

\begin{figure}[h]
    \centering
    \includegraphics[width = 0.95\textwidth]{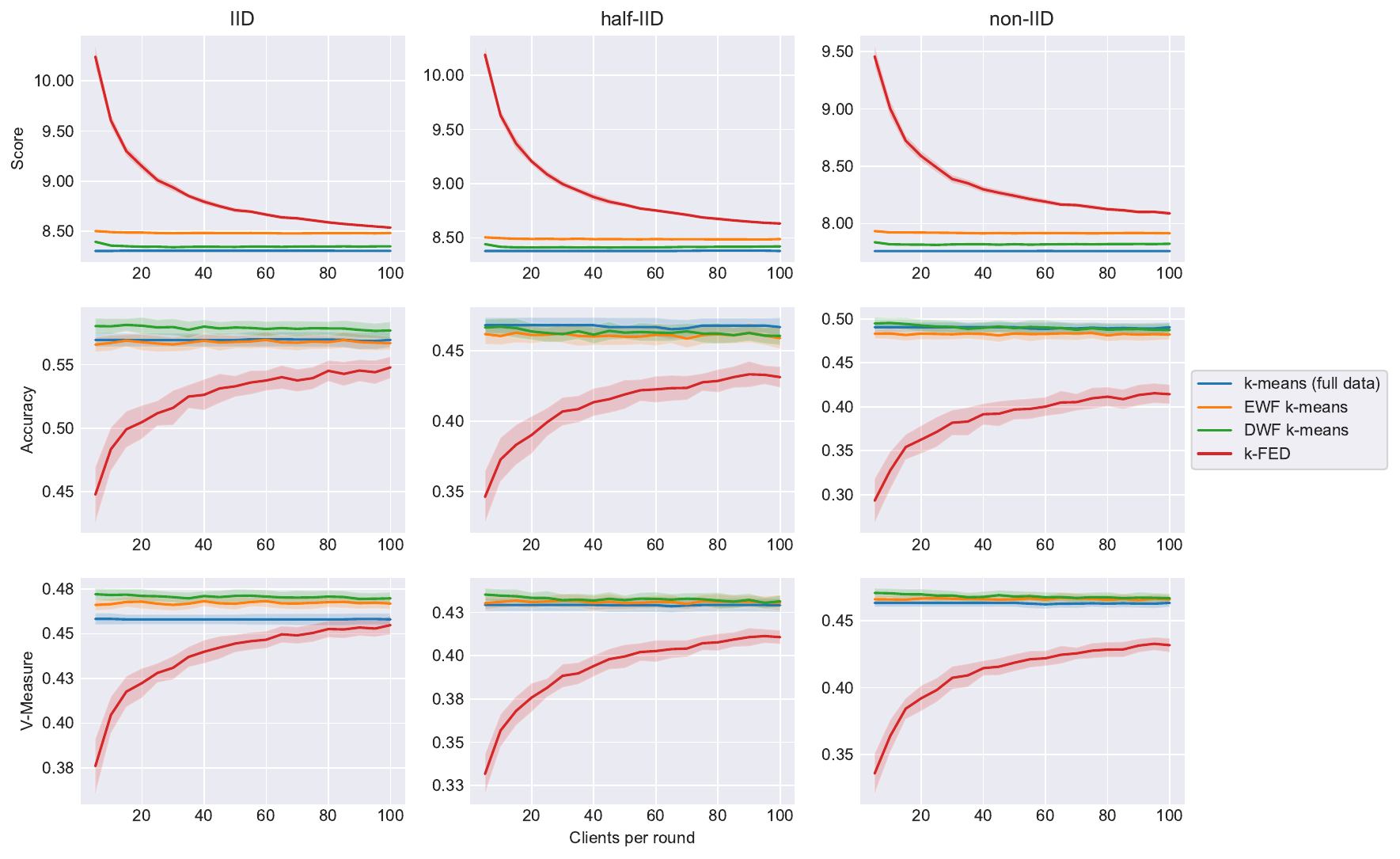}
    \caption{The results for FEMNIST. Analogous to Figure \ref{fig:experiments synth 100 all metrics}.}
    \label{fig:experiments FEMNIST 100 all metrics}
\end{figure}

In contrast to the previous results, the outcome of the FEMNIST experiments was quite different, as shown in Figure \ref{fig:experiments FEMNIST 100 all metrics}.
This is because the data distribution was generated by LEAF and is therefore distinct from those of the synthetic dataset and MNIST. As Figure \ref{fig:distribution femnist client 1} shows, the IID data is not really IID on the clients, but IID on the integers and IID on the letters. The sizes of the data on the clients also vary in the IID case for the FEMNIST data, so are also not IID in that sense. This may partially explain why there were no large differences in the results between the three distributions here.

Another difference from the previous results is that the metrics for \texttt{EWF k-means} and \texttt{DWF k-means} are relatively constant and therefore more stable, regardless of the number of clients participating in each global training round. There was only a small increase in the score when the proportion of participating clients was very low. Both \texttt{EWF k-means} and \texttt{DWF k-means} performed better than \texttt{k-FED} according to all metrics, and \texttt{DWF k-means} performed slightly better than \texttt{EWF k-means}. 
Further, we observed that \texttt{EWF k-means} and \texttt{DWF k-means} sometimes even outperform classical k-means: namely for accuracy and v-measure in the IID case, and for v-measure in the half-IID case. Since the algorithms are designed to optimize the score, it can be the case that some solutions are slightly worse in the score but better in some other metrics, as we see here.

The Tables \ref{tab:results score for all 100 clients}, \ref{tab:results accuracy for all 100 clients} and \ref{tab:results v-measure for all 100 clients} in appendix \ref{appendix:results} provide the numerical results of the experiments for all three metrics for the case where all 100 clients participated during training, i.e. for $n_{\text{clients}}=100$.

In summary, considering performance over a variable percentage of participating clients is important to align with real-world federated learning scenarios. The results of our experiments affirm that our proposed \texttt{DWF k-means} is a competitive method across varying levels of data heterogeneity and client participation. Additionally, the option to assign custom weights in \texttt{DWF k-means} makes it adaptable to different types of data heterogeneity. Even in the case where all 100 clients participate in each global training round, \texttt{EWF k-means} and \texttt{DWF k-means} perform at least as well as, or sometimes even outperform \texttt{k-FED}.

%%%%%%%%%%%%%%%%%%%%%%%%%%%%%%%%%%%%%%%%%%%%%%%%%%%%%%%%%%%%%%%%%%%%%%%%%%%%%%%%%%%%
%%%%%%%%%%%%%%%%%%%%%%%%%%%%%%%%%%%%%%%%%%%%%%%%%%%%%%%%%%%%%%%%%%%%%%%%%%%%%%%%%%%%
%%%%%%%%%%%%%%%%%%%%%%%%%%%%%%%%%%%%%%%%%%%%%%%%%%%%%%%%%%%%%%%%%%%%%%%%%%%%%%%%%%%%
\FloatBarrier
\section{Conclusion and Future Work} 
\label{sec:chapter_conclusion}

This research introduces a federated clustering technique called \texttt{Dynamically Weighted Federated K-means (DWF k-means)}, designed to mitigate the issues associated with federated learning on heterogeneous data. Specifically, it addresses the challenges by applying weights that are dynamically calculated per cluster, per client, and per round, to each global model update. This allows the algorithm to place more importance on certain updates than others, to account for the data heterogeneity when forming clusters. Consequently, our technique is more general and more powerful than the commonly used constant averaging in federated learning. 

The comparative analysis with a classical k-means baseline, \texttt{k-FED} and \texttt{EWF k-means} demonstrates the advantages of this method. We motivated \texttt{DWF k-means} as a generalization of the distributed version of k-means, resulting in the same centroids if only one k-means step is performed locally. Additionally, our experimental results corroborate that our method can match the performance of the centralized classical k-means baseline, and outperform existing federated clustering methods in realistic scenarios. 
 
The method by which the model update weights are calculated is deliberately left generic (with \texttt{WF k-means}) and we demonstrate one specific possibility in this work (with \texttt{DWF k-means}), calculating the weights based on the size of local clusters at each round. The exploration of other alternatives for weight calculation is left for future work. This could include weightings that incorporate other local cluster properties, such as the cluster score, homogeneity, or completeness. Additionally, the weights could be chosen to target certain improvements, such as the selection of weights for better accuracy, convergence, privacy, or fairness.

Since it was out of the scope of this work to provide any privacy guarantees for \texttt{DWF k-means}, this is another area for future research, as mentioned in Section \ref{sec:chapter_privacy}. In addition, to quantifying the information leakage, one could consider improving the privacy of \texttt{DWF k-means} by applying techniques such as differential privacy, secure aggregation, and homomorphic encryption methods.

\section*{Funding}
This research was funded by the Federal Ministry for Economic Affairs and Climate Action  (BMWK, formerly BMWi), grant number 01MK20013A.
\section*{Code Availability}
The code used in this work is available here \url{https://github.com/PatrickITWM/DWF-k-means}.
\section*{Acknowledgments}
This research was funded by the Federal Ministry for Economic Affairs and Climate Action  (BMWK, formerly BMWi), grant number 01MK20013A. We would like to thank our colleagues Phuc Truong and Stefanie Grimm for their valuable input and feedback.

%%%%%%%%%%%%%%%%%%%%%%%%%%%%%%%%%%%%%%%%%%%%%%%%%%%%%%%%%%%%%%%%%%%%%%%%%%%%%%%%%%%%
%%%%%%%%%%%%%%%%%%%%%%%%%%%%%%%%%%%%%%%%%%%%%%%%%%%%%%%%%%%%%%%%%%%%%%%%%%%%%%%%%%%%
\FloatBarrier
%Bibliography
\bibliographystyle{unsrt}  
\bibliography{references}  

\begin{thebibliography}{10}

\bibitem{mcmahan2017communication}
Brendan McMahan, Eider Moore, Daniel Ramage, Seth Hampson, and Blaise~Aguera y~Arcas.
\newblock Communication-efficient learning of deep networks from decentralized data.
\newblock In {\em Artificial intelligence and statistics}, pages 1273--1282. PMLR, 2017.

\bibitem{rai2010survey}
Pradeep Rai and Shubha Singh.
\newblock A survey of clustering techniques.
\newblock {\em International Journal of Computer Applications}, 7(12):1--5, 2010.

\bibitem{dennis2021heterogeneity}
Don~Kurian Dennis, Tian Li, and Virginia Smith.
\newblock Heterogeneity for the win: One-shot federated clustering.
\newblock In {\em International Conference on Machine Learning}, pages 2611--2620. PMLR, 2021.

\bibitem{Wang2020FederatedCV}
Shuai Wang and Tsung-Hui Chang.
\newblock Federated clustering via matrix factorization models: From model averaging to gradient sharing.
\newblock {\em ArXiv}, abs/2002.04930, 2020.

\bibitem{Li2022DifferentiallyPV}
Zitao Li, Tianhao Wang, and Ninghui Li.
\newblock Differentially private vertical federated clustering.
\newblock {\em Proc. VLDB Endow.}, 16:1277--1290, 2022.

\bibitem{Wang2023FederatedFK}
Yi~Wang, Jiahao Ma, Ning Gao, Qingsong Wen, Liang Sun, and Hongye Guo.
\newblock Federated fuzzy k-means for privacy-preserving behavior analysis in smart grids.
\newblock {\em Applied Energy}, 2023.

\bibitem{li2020federated}
Tian Li, Anit~Kumar Sahu, Ameet Talwalkar, and Virginia Smith.
\newblock Federated learning: Challenges, methods, and future directions.
\newblock {\em IEEE signal processing magazine}, 37(3):50--60, 2020.

\bibitem{yang2021characterizing}
Chengxu Yang, Qipeng Wang, Mengwei Xu, Zhenpeng Chen, Kaigui Bian, Yunxin Liu, and Xuanzhe Liu.
\newblock Characterizing impacts of heterogeneity in federated learning upon large-scale smartphone data.
\newblock In {\em Proceedings of the Web Conference 2021}, pages 935--946, 2021.

\bibitem{abdelmoniem2023comprehensive}
Ahmed~M Abdelmoniem, Chen-Yu Ho, Pantelis Papageorgiou, and Marco Canini.
\newblock A comprehensive empirical study of heterogeneity in federated learning.
\newblock {\em IEEE Internet of Things Journal}, 2023.

\bibitem{lloyd1982least}
Stuart Lloyd.
\newblock Least squares quantization in pcm.
\newblock {\em IEEE transactions on information theory}, 28(2):129--137, 1982.

\bibitem{murtagh2011ward}
Fionn Murtagh and Pierre Legendre.
\newblock Ward's hierarchical clustering method: clustering criterion and agglomerative algorithm.
\newblock {\em arXiv preprint arXiv:1111.6285}, 2011.

\bibitem{murtagh2012algorithms}
Fionn Murtagh and Pedro Contreras.
\newblock Algorithms for hierarchical clustering: an overview.
\newblock {\em Wiley Interdisciplinary Reviews: Data Mining and Knowledge Discovery}, 2(1):86--97, 2012.

\bibitem{ester1996density}
Martin Ester, Hans-Peter Kriegel, J{\"o}rg Sander, Xiaowei Xu, et~al.
\newblock A density-based algorithm for discovering clusters in large spatial databases with noise.
\newblock In {\em kdd}, volume~96, pages 226--231, 1996.

\bibitem{banfield1993model}
Jeffrey~D Banfield and Adrian~E Raftery.
\newblock Model-based gaussian and non-gaussian clustering.
\newblock {\em Biometrics}, pages 803--821, 1993.

\bibitem{fraley2002model}
Chris Fraley and Adrian~E Raftery.
\newblock Model-based clustering, discriminant analysis, and density estimation.
\newblock {\em Journal of the American statistical Association}, 97(458):611--631, 2002.

\bibitem{ghosh2019robust}
Avishek Ghosh, Justin Hong, Dong Yin, and Kannan Ramchandran.
\newblock Robust federated learning in a heterogeneous environment.
\newblock {\em arXiv preprint arXiv:1906.06629}, 2019.

\bibitem{sattler2020clustered}
Felix Sattler, Klaus-Robert M{\"u}ller, and Wojciech Samek.
\newblock Clustered federated learning: Model-agnostic distributed multitask optimization under privacy constraints.
\newblock {\em IEEE transactions on neural networks and learning systems}, 32(8):3710--3722, 2020.

\bibitem{ghosh2020efficient}
Avishek Ghosh, Jichan Chung, Dong Yin, and Kannan Ramchandran.
\newblock An efficient framework for clustered federated learning.
\newblock {\em Advances in Neural Information Processing Systems}, 33:19586--19597, 2020.

\bibitem{briggs2020federated}
Christopher Briggs, Zhong Fan, and Peter Andras.
\newblock Federated learning with hierarchical clustering of local updates to improve training on non-iid data.
\newblock In {\em 2020 International Joint Conference on Neural Networks (IJCNN)}, pages 1--9. IEEE, 2020.

\bibitem{Kim2020DynamicCI}
Yeongwoo Kim, Ezeddin~Al Hakim, Johan Haraldson, Henrik Eriksson, Jos{\'e} Mairton~B. da~Silva, and C.~Fischione.
\newblock Dynamic clustering in federated learning.
\newblock {\em ICC 2021 - IEEE International Conference on Communications}, pages 1--6, 2020.

\bibitem{Xie2020MultiCenterFL}
Ming Xie, Guodong Long, Tao Shen, Tianyi Zhou, Xianzhi Wang, and Jing Jiang.
\newblock Multi-center federated learning.
\newblock {\em ArXiv}, abs/2108.08647, 2020.

\bibitem{kumar2020federated}
Hemant~H Kumar, VR~Karthik, and Mydhili~K Nair.
\newblock Federated k-means clustering: A novel edge ai based approach for privacy preservation.
\newblock In {\em 2020 IEEE International Conference on Cloud Computing in Emerging Markets (CCEM)}, pages 52--56. IEEE, 2020.

\bibitem{pedrycz2021federated}
Witold Pedrycz.
\newblock Federated fcm: clustering under privacy requirements.
\newblock {\em IEEE Transactions on Fuzzy Systems}, 30(8):3384--3388, 2021.

\bibitem{stallmann2022towards}
Morris Stallmann and Anna Wilbik.
\newblock Towards federated clustering: A federated fuzzy $ c $-means algorithm (ffcm).
\newblock {\em arXiv preprint arXiv:2201.07316}, 2022.

\bibitem{Garst2023FederatedKC}
Swier Garst and Marcel Reinders.
\newblock Federated k-means clustering.
\newblock {\em ArXiv}, abs/2310.01195, 2023.

\bibitem{Ding2016KMeansCW}
Hu~Ding, Yu~Liu, Lingxiao Huang, and J.~Li.
\newblock K-means clustering with distributed dimensions.
\newblock In {\em International Conference on Machine Learning}, 2016.

\bibitem{dhillon2002data}
Inderjit~S Dhillon and Dharmendra~S Modha.
\newblock A data-clustering algorithm on distributed memory multiprocessors.
\newblock In {\em Large-scale parallel data mining}, pages 245--260. Springer, 2002.

\bibitem{xu2002fast}
Xiaowei Xu, Jochen J{\"a}ger, and Hans-Peter Kriegel.
\newblock A fast parallel clustering algorithm for large spatial databases.
\newblock {\em High Performance Data Mining: Scaling Algorithms, Applications and Systems}, pages 263--290, 2002.

\bibitem{tasoulis2004unsupervised}
Dimitris~K Tasoulis and Michael~N Vrahatis.
\newblock Unsupervised distributed clustering.
\newblock In {\em Parallel and distributed computing and networks}, pages 347--351. Citeseer, 2004.

\bibitem{su2001modified}
Mu-Chun Su and Chien-Hsing Chou.
\newblock A modified version of the k-means algorithm with a distance based on cluster symmetry.
\newblock {\em IEEE Transactions on pattern analysis and machine intelligence}, 23(6):674--680, 2001.

\bibitem{manju2018ac}
Vethamuthu~Nesamony Manju and Alfred Lenin~Fred.
\newblock Ac coefficient and k-means cuckoo optimisation algorithm-based segmentation and compression of compound images.
\newblock {\em IET Image Processing}, 12(2):218--225, 2018.

\bibitem{bai2018ensemble}
Liang Bai, Jiye Liang, and Yike Guo.
\newblock An ensemble clusterer of multiple fuzzy $ k $-means clusterings to recognize arbitrarily shaped clusters.
\newblock {\em IEEE Transactions on Fuzzy Systems}, 26(6):3524--3533, 2018.

\bibitem{chen2016revisiting}
Jianmin Chen, Xinghao Pan, Rajat Monga, Samy Bengio, and Rafal Jozefowicz.
\newblock Revisiting distributed synchronous sgd.
\newblock {\em arXiv preprint arXiv:1604.00981}, 2016.

\bibitem{scikit-learn}
F.~Pedregosa, G.~Varoquaux, A.~Gramfort, V.~Michel, B.~Thirion, O.~Grisel, M.~Blondel, P.~Prettenhofer, R.~Weiss, V.~Dubourg, J.~Vanderplas, A.~Passos, D.~Cournapeau, M.~Brucher, M.~Perrot, and E.~Duchesnay.
\newblock Scikit-learn: Machine learning in {P}ython.
\newblock {\em Journal of Machine Learning Research}, 12:2825--2830, 2011.

\bibitem{yang2018applied}
Timothy Yang, Galen Andrew, Hubert Eichner, Haicheng Sun, Wei Li, Nicholas Kong, Daniel Ramage, and Françoise Beaufays.
\newblock Applied federated learning: Improving google keyboard query suggestions, 2018.

\bibitem{chen2019federated}
Mingqing Chen, Rajiv Mathews, Tom Ouyang, and Françoise Beaufays.
\newblock Federated learning of out-of-vocabulary words, 2019.

\bibitem{Apple2019designforprivacy}
Apple Inc.
\newblock Designing for privacy - (video and slide deck), Jun 2019.

\bibitem{leroy2019federated}
David Leroy, Alice Coucke, Thibaut Lavril, Thibault Gisselbrecht, and Joseph Dureau.
\newblock Federated learning for keyword spotting.
\newblock In {\em ICASSP 2019-2019 IEEE international conference on acoustics, speech and signal processing (ICASSP)}, pages 6341--6345. IEEE, 2019.

\bibitem{EUCORDISDrug}
EU~CORDIS.
\newblock Machine learning ledger orchestration for drug discovery.

\bibitem{erlingsson2014rappor}
{\'U}lfar Erlingsson, Vasyl Pihur, and Aleksandra Korolova.
\newblock Rappor: Randomized aggregatable privacy-preserving ordinal response.
\newblock In {\em Proceedings of the 2014 ACM SIGSAC conference on computer and communications security}, pages 1054--1067, 2014.

\bibitem{al2019privacy}
Mohammad Al-Rubaie and J~Morris Chang.
\newblock Privacy-preserving machine learning: Threats and solutions.
\newblock {\em IEEE Security \& Privacy}, 17(2):49--58, 2019.

\bibitem{hu2022membership}
Hongsheng Hu, Zoran Salcic, Lichao Sun, Gillian Dobbie, Philip~S Yu, and Xuyun Zhang.
\newblock Membership inference attacks on machine learning: A survey.
\newblock {\em ACM Computing Surveys (CSUR)}, 54(11s):1--37, 2022.

\bibitem{rigaki2020survey}
Maria Rigaki and Sebastian Garcia.
\newblock A survey of privacy attacks in machine learning.
\newblock {\em arXiv preprint arXiv:2007.07646}, 2020.

\bibitem{wang2019beyond}
Zhibo Wang, Mengkai Song, Zhifei Zhang, Yang Song, Qian Wang, and Hairong Qi.
\newblock Beyond inferring class representatives: User-level privacy leakage from federated learning.
\newblock In {\em IEEE INFOCOM 2019-IEEE conference on computer communications}, pages 2512--2520. IEEE, 2019.

\bibitem{mothukuri2021survey}
Viraaji Mothukuri, Reza~M Parizi, Seyedamin Pouriyeh, Yan Huang, Ali Dehghantanha, and Gautam Srivastava.
\newblock A survey on security and privacy of federated learning.
\newblock {\em Future Generation Computer Systems}, 115:619--640, 2021.

\bibitem{bagdasaryan2020backdoor}
Eugene Bagdasaryan, Andreas Veit, Yiqing Hua, Deborah Estrin, and Vitaly Shmatikov.
\newblock How to backdoor federated learning.
\newblock In {\em International conference on artificial intelligence and statistics}, pages 2938--2948. PMLR, 2020.

\bibitem{sun2019can}
Ziteng Sun, Peter Kairouz, Ananda~Theertha Suresh, and H~Brendan McMahan.
\newblock Can you really backdoor federated learning?
\newblock {\em arXiv preprint arXiv:1911.07963}, 2019.

\bibitem{wang2020attack}
Hongyi Wang, Kartik Sreenivasan, Shashank Rajput, Harit Vishwakarma, Saurabh Agarwal, Jy-yong Sohn, Kangwook Lee, and Dimitris Papailiopoulos.
\newblock Attack of the tails: Yes, you really can backdoor federated learning.
\newblock {\em Advances in Neural Information Processing Systems}, 33:16070--16084, 2020.

\bibitem{dwork2006calibrating}
Cynthia Dwork, Frank McSherry, Kobbi Nissim, and Adam Smith.
\newblock Calibrating noise to sensitivity in private data analysis.
\newblock In {\em Theory of Cryptography: Third Theory of Cryptography Conference, TCC 2006, New York, NY, USA, March 4-7, 2006. Proceedings 3}, pages 265--284. Springer, 2006.

\bibitem{dwork2008differential}
Cynthia Dwork.
\newblock Differential privacy: A survey of results.
\newblock In {\em International conference on theory and applications of models of computation}, pages 1--19. Springer, 2008.

\bibitem{dwork2014algorithmic}
Cynthia Dwork, Aaron Roth, et~al.
\newblock The algorithmic foundations of differential privacy.
\newblock {\em Foundations and Trends{\textregistered} in Theoretical Computer Science}, 9(3--4):211--407, 2014.

\bibitem{wei2020federated}
Kang Wei, Jun Li, Ming Ding, Chuan Ma, Howard~H Yang, Farhad Farokhi, Shi Jin, Tony~QS Quek, and H~Vincent Poor.
\newblock Federated learning with differential privacy: Algorithms and performance analysis.
\newblock {\em IEEE Transactions on Information Forensics and Security}, 15:3454--3469, 2020.

\bibitem{yao1986generate}
Andrew Chi-Chih Yao.
\newblock How to generate and exchange secrets.
\newblock In {\em 27th annual symposium on foundations of computer science (Sfcs 1986)}, pages 162--167. IEEE, 1986.

\bibitem{mohassel2017secureml}
Payman Mohassel and Yupeng Zhang.
\newblock Secureml: A system for scalable privacy-preserving machine learning.
\newblock In {\em 2017 IEEE symposium on security and privacy (SP)}, pages 19--38. IEEE, 2017.

\bibitem{gentry2009fully}
Craig Gentry.
\newblock Fully homomorphic encryption using ideal lattices.
\newblock In {\em Proceedings of the forty-first annual ACM symposium on Theory of computing}, pages 169--178, 2009.

\bibitem{zhang2020batchcrypt}
Chengliang Zhang, Suyi Li, Junzhe Xia, Wei Wang, Feng Yan, and Yang Liu.
\newblock $\{$BatchCrypt$\}$: Efficient homomorphic encryption for $\{$Cross-Silo$\}$ federated learning.
\newblock In {\em 2020 USENIX annual technical conference (USENIX ATC 20)}, pages 493--506, 2020.

\bibitem{bonawitz2017practical}
Keith Bonawitz, Vladimir Ivanov, Ben Kreuter, Antonio Marcedone, H~Brendan McMahan, Sarvar Patel, Daniel Ramage, Aaron Segal, and Karn Seth.
\newblock Practical secure aggregation for privacy-preserving machine learning.
\newblock In {\em proceedings of the 2017 ACM SIGSAC Conference on Computer and Communications Security}, pages 1175--1191, 2017.

\bibitem{bell2020secure}
James~Henry Bell, Kallista~A Bonawitz, Adri{\`a} Gasc{\'o}n, Tancr{\`e}de Lepoint, and Mariana Raykova.
\newblock Secure single-server aggregation with (poly) logarithmic overhead.
\newblock In {\em Proceedings of the 2020 ACM SIGSAC Conference on Computer and Communications Security}, pages 1253--1269, 2020.

\bibitem{lecun2010mnist}
Yann LeCun, Corinna Cortes, and CJ~Burges.
\newblock Mnist handwritten digit database.
\newblock {\em ATT Labs [Online]. Available: http://yann.lecun.com/exdb/mnist}, 2, 2010.

\bibitem{caldas2018leaf}
Sebastian Caldas, Sai Meher~Karthik Duddu, Peter Wu, Tian Li, Jakub Kone{\v{c}}n{\`y}, H~Brendan McMahan, Virginia Smith, and Ameet Talwalkar.
\newblock Leaf: A benchmark for federated settings.
\newblock {\em arXiv preprint arXiv:1812.01097}, 2018.

\bibitem{rosenberg2007vmeassure}
Andrew Rosenberg and Julia Hirschberg.
\newblock V-measure: A conditional entropy-based external cluster evaluation measure.
\newblock In {\em Proceedings of the 2007 joint conference on empirical methods in natural language processing and computational natural language learning (EMNLP-CoNLL)}, pages 410--420, 2007.

\end{thebibliography}

%%%%%%%%%%%%%%%%%%%%%%%%%%%%%%%%%%%%%%%%%%%%%%%%%%%%%%%%%%%%%%%%%%%%%%%%%%%%%%%%%%%%
%%%%%%%%%%%%%%%%%%%%%%%%%%%%%%%%%%%%%%%%%%%%%%%%%%%%%%%%%%%%%%%%%%%%%%%%%%%%%%%%%%%%

%%%%%%%%%%%%%%%%%%%%%%%%%%%%%%%%%%%%%%%%%%
%% Optional
\appendix
\section[\appendixname~\thesection]{}
\label{appendix:algorithms}
% ---------------------------------------------------
Algorithm \ref{alg:update_local_centroids}, \ref{alg:aggregate_centroids} and \ref{alg:request_score} complement Algorithm \ref{alg:dwfkm} and specify the single intermediate steps. Note that the weights computed in Algorithm \ref{alg:update_local_centroids} are the ones for \texttt{DWF k-means} - if other weights are used line 3 should be modified. For a matrix $\mathbf{M} \in \R^{m \times n}$, we denote the j-th column by $\mathbf{M}_{\cdot, j}$.
\begin{algorithm}[h]
\caption{LocalUpdate}
\label{alg:update_local_centroids}
\SetKwInOut{Input}{Input}\SetKwInOut{Output}{Output}
\SetKwProg{ForAllInParallel}{foreach}{ do in parallel}{end} 
\Input{Local data $X_{i}$, 
global centroids $\textbf{C}$, 
number of maximal local iterations $\text{max}_{\text{local}}$,
tolerance $\epsilon >0$.
 }
\Output{The locally updated centroids $\textbf{C}_{i}  \in \mathbb{R}^{m  \times k}$ and the assigned weights $\omega_{i} \in \mathbb{R}^{k}$.}
 $\textbf{C}_i \leftarrow$ Perform classical k-means with $\textbf{C}$ as initial centroids, and arguments $X_{i}$, $k$, $\text{max}_{\text{local}}$ and $\epsilon$\;
\For{$i =$ 1,..., k}{
$\omega_{i, j} \leftarrow |\{x \in X_{i}| \ x\text{ is assigned to centroid } \textbf{c}_j\}|$
}
\textbf{return} $\textbf{C}_i$, $\omega_i$\;
\end{algorithm}
% ---------------------------------------------------
\begin{algorithm}[h!]
\caption{Aggregate}
\label{alg:aggregate_centroids}
\SetKwInOut{Input}{Input}\SetKwInOut{Output}{Output}
\SetKwProg{ForAllInParallel}{foreach}{ do in parallel}{end} 
\Input{Local centroids $(\textbf{C}_{i})_{i \in I_t}$, 
 weights $(\omega_{i})_{i \in I_t}$,
 last global centroids $\textbf{C}$, 
 penultimate centroids $\textbf{C}_{\text{old}}$, 
 learning rate $\eta$, 
 momentum $\mu$.
 }
\Output{Aggregated centroids $\textbf{C}_{\text{new}} \in \R^{m\times k}$, where $m$ is the dimension of the data.}
\textbf{s} $\leftarrow \sum_{i \in I_t} \omega_i \in \mathbb{R}^k$\;
    \ForAll{$j= 1,..., k$}{
    \tcc{Catch the case where all weights are zero}
        \uIf{$\mathbf{s}_j = 0$}{
            $\lambda_j \leftarrow \frac{1}{|I_t|}$\;
        }
        \uElse{$\lambda_j \leftarrow \omega_j$\;}
        $(\textbf{D})_{\cdot, j}\leftarrow \frac{1}{\textbf{s}_s} \sum_{i \in I_t} \lambda_j \cdot (\textbf{C}_{i})_{\cdot, j}$\;
        $\textbf{C}_{\text{new}}\leftarrow \textbf{C} + \eta \cdot (\textbf{D}-\textbf{C}) + \mu \cdot (\textbf{C}-\textbf{C}_{\text{old}})$;
    }
     
\textbf{return} $\textbf{C}_{\text{new}}$\;
\end{algorithm}
% ---------------------------------------------------
\begin{algorithm}[h!]
\caption{RequestScore}
\label{alg:request_score}
\SetKwInOut{Input}{Input}\SetKwInOut{Output}{Output}
\SetKwProg{ForAllInParallel}{foreach}{ do in parallel}{end} 
\Input{
 Local data $(X_i)_{i \in [N]}$,
 centroids $\textbf{C}$.
 }
\Output{A score $s \in \mathbb{R}$.}
s $\leftarrow 0$\;
Choose non empty $B \subseteq [N]$\;
\tcc{If a client does not respond in time, it can be considered as not in $B$.}
\ForAllInParallel{$i \in B$}{
    \tcc{compute the classical k-means score.}
    $s_i \leftarrow$ score($X_i$, $\textbf{C}$)\;
}
\textbf{return} $\frac{\sum_{i \in B} s_i}{|B|}$\;
\end{algorithm}
% ---------------------------------------------------
% ---------------------------------------------------
% ---------------------------------------------------

%%%%%%%%%%%%%%%%%%%%%%%%%%%%%%%%%%%%%%%%%%%%%%%%%%%%%%%%%%%%%%%%%%%%%%%%%%%%%%%%%%%%
\FloatBarrier
\section[\appendixname~\thesection]{}
\label{appendix:results}
\begin{figure}[h]
    \centering
    \includegraphics[width = 0.95\textwidth]{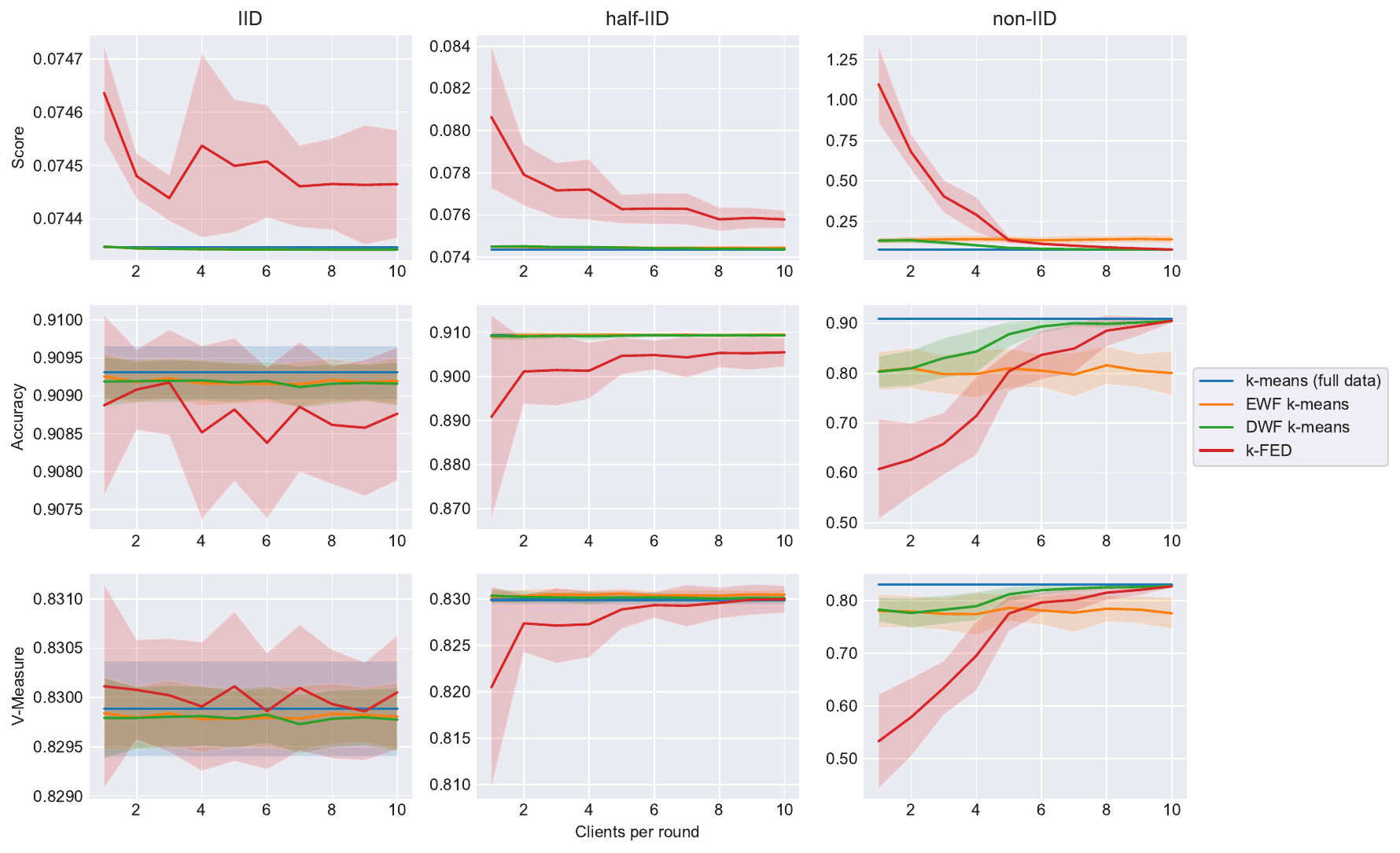}
    \caption{The results for the synthetic data, where the data is distributed to only 10 clients. To remove results where the algorithms became stuck at some local minima, for each of the values for $n_{\text{clients}}$, from 100 runs we only kept the 50 runs with the lowest score. The solid line is the mean $m$ of the calculated metrics of the 50 runs (dependent on $n_{\text{clients}}$), the area around it is the mean $\pm$ the standard deviation of these 50 runs.}
    \label{fig:experiments synth 10 all metrics}
\end{figure}

\begin{figure}[h]
    \centering
    \includegraphics[width = 0.95\textwidth]{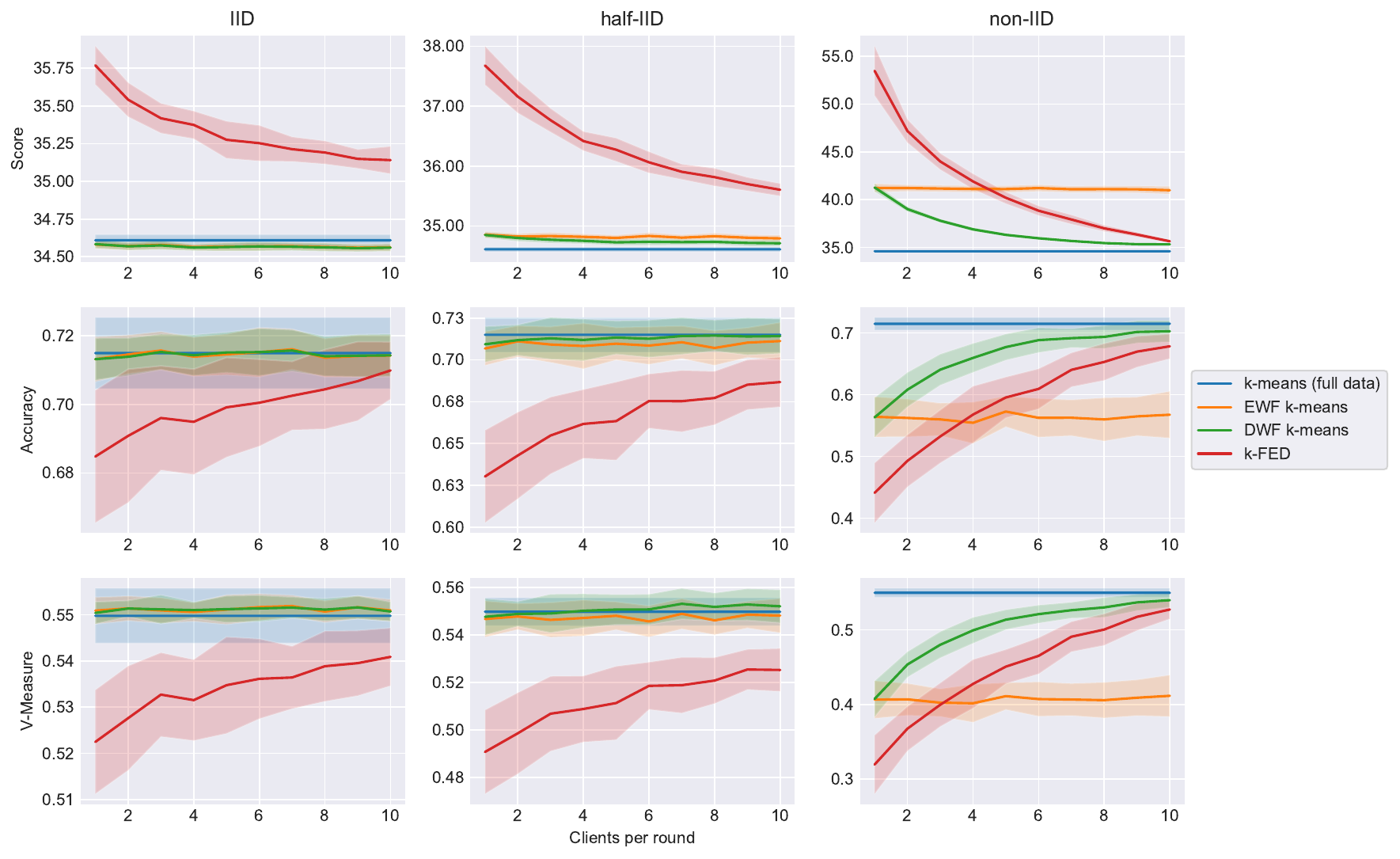}
    \caption{The results for MNIST, where the data is distributed to only 10 clients, analogous to Figure \ref{fig:experiments synth 10 all metrics}.}
    \label{fig:experiments MNIST 10 all metrics}
\end{figure}

For the synthetic dataset and MNIST, we also ran the experiments with the same settings as in section \ref{subsec:parameters}, but with the data distributed to 10 instead of 100 clients. We iterated over $n_{\text{clients}} = 1,2,...,10$ instead of $n_{\text{clients}} = 5,10,...,100$. The results can be found in Figures \ref{fig:experiments synth 10 all metrics} and \ref{fig:experiments MNIST 10 all metrics}.
The results are consistent with those from the experiments with 100 clients.

\FloatBarrier
\begin{table}[H] 
\begin{tabular}{lllrrrr}
\toprule
 &  &  & k-means & EWF k-means & DWF k-means & k-FED \\
Data & Dist. & Value &  &  &  &  \\
\midrule
\multirow[t]{9}{*}{Synthetic data} & \multirow[t]{3}{*}{IID} & mean & 0.0780 & 0.0743 & 0.0743 & 0.0759 \\
 &  & min & 0.0743 & 0.0743 & 0.0743 & 0.0746 \\
 &  & std & 0.0066 & 0.0000 & 0.0000 & 0.0039 \\
\cline{2-7}
 & \multirow[t]{3}{*}{half-IID} & mean & 0.0780 & 0.0744 & 0.0744 & 0.0786 \\
 &  & min & 0.0743 & 0.0744 & 0.0743 & 0.0746 \\
 &  & std & 0.0066 & 0.0000 & 0.0000 & 0.0066 \\
\cline{2-7}
 & \multirow[t]{3}{*}{non-IID} & mean & 0.0780 & 0.0772 & 0.0744 & 0.0817 \\
 &  & min & 0.0743 & 0.0745 & 0.0743 & 0.0755 \\
 &  & std & 0.0066 & 0.0020 & 0.0000 & 0.0071 \\
\cline{1-7} \cline{2-7}
\multirow[t]{9}{*}{MNIST} & \multirow[t]{3}{*}{IID} & mean & 34.6892 & 34.7191 & 34.6930 & 35.7251 \\
 &  & min & 34.5299 & 34.5549 & 34.5326 & 35.3449 \\
 &  & std & 0.0987 & 0.1094 & 0.1113 & 0.2272 \\
\cline{2-7}
 & \multirow[t]{3}{*}{half-IID} & mean & 34.6892 & 34.9325 & 34.7729 & 35.7704 \\
 &  & min & 34.5299 & 34.6773 & 34.5480 & 35.2623 \\
 &  & std & 0.0987 & 0.1695 & 0.1747 & 0.2733 \\
\cline{2-7}
 & \multirow[t]{3}{*}{non-IID} & mean & 34.6892 & 38.6028 & 34.7879 & 35.4158 \\
 &  & min & 34.5299 & 37.8510 & 34.6328 & 35.0402 \\
 &  & std & 0.0987 & 0.3471 & 0.0870 & 0.2038 \\
\cline{1-7} \cline{2-7}
\multirow[t]{9}{*}{FEMNIST} & \multirow[t]{3}{*}{IID} & mean & 8.3140 & 8.4945 & 8.3624 & 8.5574 \\
 &  & min & 8.2876 & 8.4694 & 8.3323 & 8.4907 \\
 &  & std & 0.0111 & 0.0131 & 0.0140 & 0.0258 \\
\cline{2-7}
 & \multirow[t]{3}{*}{half-IID} & mean & 8.3871 & 8.4991 & 8.4309 & 8.6519 \\
 &  & min & 8.3691 & 8.4736 & 8.4046 & 8.5996 \\
 &  & std & 0.0101 & 0.0125 & 0.0121 & 0.0245 \\
\cline{2-7}
 & \multirow[t]{3}{*}{non-IID} & mean & 7.7661 & 7.9266 & 7.8330 & 8.1102 \\
 &  & min & 7.7458 & 7.8939 & 7.8070 & 8.0470 \\
 &  & std & 0.0075 & 0.0148 & 0.0113 & 0.0288 \\
\cline{1-7} \cline{2-7}
\bottomrule
\end{tabular}
\caption{This table presents the mean, minimum and standard deviation of the score for 100 runs per configuration and algorithm, where the data was distributed across 100 clients and all clients participated in each training round, i.e. where $n_{\text{clients}}=100$.}
\label{tab:results score for all 100 clients}
\end{table}

%%%%%%%%%%%%%%%%%%%%%%%%%%%%%%%%%%%%%%%%%%%%%%%%%%%%%%%%%%%%%%%%
\begin{table}[H] 
\begin{tabular}{lllrrrr}
\toprule
 &  &  & k-means & EWF k-means & DWF k-means & k-FED \\
Data & Dist. & Value &  &  &  &  \\
\midrule
\multirow[t]{9}{*}{Synthetic data} & \multirow[t]{3}{*}{IID} & mean & 0.9093 & 0.9091 & 0.9091 & 0.9077 \\
 &  & max & 0.9099 & 0.9098 & 0.9100 & 0.9108 \\
 &  & std & 0.0004 & 0.0003 & 0.0003 & 0.0015 \\
\cline{2-7}
 & \multirow[t]{3}{*}{half-IID} & mean & 0.9093 & 0.9093 & 0.9093 & 0.9083 \\
 &  & max & 0.9099 & 0.9102 & 0.9103 & 0.9108 \\
 &  & std & 0.0004 & 0.0003 & 0.0003 & 0.0019 \\
\cline{2-7}
 & \multirow[t]{3}{*}{non-IID} & mean & 0.9093 & 0.9049 & 0.9093 & 0.9062 \\
 &  & max & 0.9099 & 0.9104 & 0.9099 & 0.9091 \\
 &  & std & 0.0004 & 0.0042 & 0.0003 & 0.0023 \\
\cline{1-7} \cline{2-7}
\multirow[t]{9}{*}{MNIST} & \multirow[t]{3}{*}{IID} & mean & 0.7150 & 0.7144 & 0.7149 & 0.6789 \\
 &  & max & 0.7383 & 0.7288 & 0.7329 & 0.7067 \\
 &  & std & 0.0104 & 0.0061 & 0.0061 & 0.0130 \\
\cline{2-7}
 & \multirow[t]{3}{*}{half-IID} & mean & 0.7150 & 0.7012 & 0.7124 & 0.6692 \\
 &  & max & 0.7383 & 0.7157 & 0.7271 & 0.7182 \\
 &  & std & 0.0104 & 0.0081 & 0.0063 & 0.0156 \\
\cline{2-7}
 & \multirow[t]{3}{*}{non-IID} & mean & 0.7150 & 0.5983 & 0.7037 & 0.6861 \\
 &  & max & 0.7383 & 0.6308 & 0.7332 & 0.7122 \\
 &  & std & 0.0104 & 0.0162 & 0.0092 & 0.0124 \\
\cline{1-7} \cline{2-7}
\multirow[t]{9}{*}{FEMNIST} & \multirow[t]{3}{*}{IID} & mean & 0.5695 & 0.5669 & 0.5769 & 0.5479 \\
 &  & max & 0.5814 & 0.5763 & 0.5935 & 0.5645 \\
 &  & std & 0.0058 & 0.0053 & 0.0069 & 0.0085 \\
\cline{2-7}
 & \multirow[t]{3}{*}{half-IID} & mean & 0.4667 & 0.4589 & 0.4605 & 0.4312 \\
 &  & max & 0.4848 & 0.4774 & 0.4747 & 0.4504 \\
 &  & std & 0.0066 & 0.0075 & 0.0066 & 0.0074 \\
\cline{2-7}
 & \multirow[t]{3}{*}{non-IID} & mean & 0.4906 & 0.4823 & 0.4874 & 0.4143 \\
 &  & max & 0.5029 & 0.4937 & 0.5024 & 0.4389 \\
 &  & std & 0.0056 & 0.0060 & 0.0078 & 0.0110 \\
\cline{1-7} \cline{2-7}
\bottomrule
\end{tabular}
\caption{This table presents the mean, maximum and standard deviation of the accuracy for 100 runs per configuration and algorithm, where the data was distributed across 100 clients and all clients participated in each training round, i.e. where $n_{\text{clients}}=100$.}
\label{tab:results accuracy for all 100 clients}
\end{table}

%%%%%%%%%%%%%%%%%%%%%%%%%%%%%%%%%%%%%%%%%%%%%%%%%%%%%%%%%%%%%%%%
\begin{table}[H] 
\begin{tabular}{lllrrrr}
\toprule
 &  &  & k-means & EWF k-means & DWF k-means & k-FED \\
Data & Dist. & Value &  &  &  &  \\
\midrule
\multirow[t]{9}{*}{Synthetic data} & \multirow[t]{3}{*}{IID} & mean & 0.8299 & 0.8297 & 0.8297 & 0.8301 \\
 &  & max & 0.8309 & 0.8310 & 0.8312 & 0.8327 \\
 &  & std & 0.0005 & 0.0003 & 0.0003 & 0.0011 \\
\cline{2-7}
 & \multirow[t]{3}{*}{half-IID} & mean & 0.8299 & 0.8300 & 0.8299 & 0.8307 \\
 &  & max & 0.8309 & 0.8317 & 0.8311 & 0.8330 \\
 &  & std & 0.0005 & 0.0004 & 0.0004 & 0.0012 \\
\cline{2-7}
 & \multirow[t]{3}{*}{non-IID} & mean & 0.8299 & 0.8280 & 0.8298 & 0.8269 \\
 &  & max & 0.8309 & 0.8322 & 0.8309 & 0.8303 \\
 &  & std & 0.0005 & 0.0027 & 0.0004 & 0.0019 \\
\cline{1-7} \cline{2-7}
\multirow[t]{9}{*}{MNIST} & \multirow[t]{3}{*}{IID} & mean & 0.5498 & 0.5509 & 0.5517 & 0.5174 \\
 &  & max & 0.5596 & 0.5605 & 0.5627 & 0.5403 \\
 &  & std & 0.0059 & 0.0042 & 0.0038 & 0.0068 \\
\cline{2-7}
 & \multirow[t]{3}{*}{half-IID} & mean & 0.5498 & 0.5392 & 0.5496 & 0.5130 \\
 &  & max & 0.5596 & 0.5508 & 0.5592 & 0.5468 \\
 &  & std & 0.0059 & 0.0056 & 0.0040 & 0.0090 \\
\cline{2-7}
 & \multirow[t]{3}{*}{non-IID} & mean & 0.5498 & 0.4338 & 0.5410 & 0.5297 \\
 &  & max & 0.5596 & 0.4685 & 0.5519 & 0.5524 \\
 &  & std & 0.0059 & 0.0136 & 0.0052 & 0.0083 \\
\cline{1-7} \cline{2-7}
\multirow[t]{9}{*}{FEMNIST} & \multirow[t]{3}{*}{IID} & mean & 0.4580 & 0.4668 & 0.4697 & 0.4548 \\
 &  & max & 0.4680 & 0.4732 & 0.4787 & 0.4658 \\
 &  & std & 0.0034 & 0.0032 & 0.0034 & 0.0052 \\
\cline{2-7}
 & \multirow[t]{3}{*}{half-IID} & mean & 0.4292 & 0.4306 & 0.4314 & 0.4107 \\
 &  & max & 0.4389 & 0.4406 & 0.4424 & 0.4201 \\
 &  & std & 0.0039 & 0.0044 & 0.0036 & 0.0040 \\
\cline{2-7}
 & \multirow[t]{3}{*}{non-IID} & mean & 0.4632 & 0.4657 & 0.4667 & 0.4317 \\
 &  & max & 0.4691 & 0.4720 & 0.4745 & 0.4440 \\
 &  & std & 0.0031 & 0.0034 & 0.0035 & 0.0053 \\
\cline{1-7} \cline{2-7}
\bottomrule
\end{tabular}
\caption{This table presents the mean, maximum and standard deviation of the v-measure for 100 runs per configuration and algorithm, where the data was distributed across 100 clients and all clients participated in each training round, i.e. where $n_{\text{clients}}=100$.}
\label{tab:results v-measure for all 100 clients}
\end{table}
%%%%%%%%%%%%%%%%%%%%%%%%%%%%%%%%%%%%%%%%%%%%%%%%%%%%%%%%%%%%%%%%%%%%%%%%%%%%%%%%%%%%
%%%%%%%%%%%%%%%%%%%%%%%%%%%%%%%%%%%%%%%%%%%%%%%%%%%%%%%%%%%%%%%%%%%%%%%%%%%%%%%%%%%%

\end{document}